\documentclass[conference]{IEEEtran}
\IEEEoverridecommandlockouts
\usepackage{cite}
\usepackage[hyphens]{url}
\usepackage{graphicx}
\usepackage{algorithmic}
\usepackage{amssymb}
\usepackage{amsmath,amsfonts}
\usepackage{textcomp}
\usepackage{xcolor}
\usepackage{url}
\usepackage{adjustbox}
\usepackage{multirow}
\usepackage{enumitem}
\usepackage{bm}
\usepackage[caption=false]{subfig}
\usepackage[ruled, lined, linesnumbered, commentsnumbered, longend]{algorithm2e}
\usepackage{xcolor}
\usepackage{newfloat}
\usepackage{listings}

\usepackage{setspace}

\usepackage{amsthm}
\newtheorem{theorem}{Theorem}

\title{Demystifying Distributed Training of Graph Neural Networks for Link Prediction}

\author{\IEEEauthorblockN{Xin Huang and Chul-Ho Lee}
\IEEEauthorblockA{Department of Computer Science, Texas State University \\
\{xhuang, chulho.lee\}@txstate.edu}
}

\newcommand{\n}{SpLPG}
\begin{document}

\maketitle

\begin{abstract}
Graph neural networks (GNNs) are powerful tools for solving graph-related problems. Distributed GNN frameworks and systems enhance the scalability of GNNs and accelerate model training, yet most are optimized for node classification. Their performance on link prediction remains underexplored. This paper demystifies distributed training of GNNs for link prediction by investigating the issue of performance degradation when each worker trains a GNN on its assigned partitioned subgraph without having access to the entire graph. We discover that the main sources of the issue come from not only the information loss caused by graph partitioning but also the ways of drawing negative samples during model training. While sharing the complete graph information with each worker resolves the issue and preserves link prediction accuracy, it incurs a high communication cost. We propose \n{}, which effectively leverages graph sparsification to mitigate the issue of performance degradation at a reduced communication cost. Experiment results on several public real-world datasets demonstrate the effectiveness of \n{}, which reduces the communication overhead by up to about 80\% while mostly preserving link prediction accuracy.
\end{abstract}

\section{Introduction}

Link prediction is to infer missing or potential connections between nodes in a graph. It serves as a fundamental task in graph mining and plays a critical role in various domains, such as social network analysis~\cite{liben2003link} and knowledge graph completion~\cite{lin2015learning}, to name a few. The link prediction problem has been widely studied in the literature, and different solutions have been proposed, including classical heuristic algorithms and machine learning approaches~\cite{lu2011link, kumar2020link,wu2022graph}. In particular, graph neural networks (GNNs)~\cite{ma2021deep, hamilton2020graph, huang2022characterizing} have recently emerged as a popular tool for link prediction because of their excellent capability in learning from graph-structured data. By using the neighborhood aggregation mechanism~\cite{hamilton2020graph}, GNNs iteratively aggregate and update features or embeddings according to the structure of the underlying graph. They have shown superior performance in a wide range of graph-related tasks~\cite{zhong2021heterogeneous,zhuo2022grafics,wu2023embedding,deng2024mega}, including link prediction.

Many studies have been focused on developing new GNN architectures for link prediction and/or applying GNNs to real-world applications involving the link prediction problem. For example, You et al. generalize the GNN models that were originally proposed for node classification~\cite{kipf2016semi,velivckovic2017graph,hamilton2017inductive,xu2018powerful} to link prediction with a dot-product edge predictor~\cite{you2019position}. They also propose a new GNN architecture called PGNN that learns position-aware node embeddings to further improve prediction performance. PinSage~\cite{ying2018graph} uses the GraphSAGE model with random walks to predict links in recommender systems. R-GCN~\cite{schlichtkrull2018modeling} is developed as an extension of graph convolutional networks (GCNs) to handle multi-relational data for knowledge graph completion. These studies assume that GNN models are trained in a centralized manner.

To handle large graphs efficiently and accelerate model training effectively, several distributed GNN training systems and frameworks~\cite{FeyLenssen2019,zheng2020distdgl,zheng2022distributed,md2021distgnn,liu2023bgl,wan2023adaptive} have been proposed and developed recently. They typically first partition the graphs into small subgraphs and then distribute the partitioned subgraphs across a cluster of workers, e.g., GPUs or machines, to enable model training in parallel. While these systems and frameworks significantly enhances scalability and computational efficiency of GNNs on large graphs, most of them are primarily designed and optimized for node classification. As mainstream GNN frameworks, DistDGL~\cite{zheng2020distdgl,zheng2022distributed} and PyG~\cite{FeyLenssen2019} support distributed GNN training for link prediction. However, due to the lack of sufficient information and complete official examples, their performance is still \emph{not} well-studied. In other words, GNNs for link prediction under distributed settings still remains underexplored.

Recently, Zhu et al. reveal the problem of a performance drop in distributed GNN training for link prediction when GNN models are trained on each partitioned subgraph in each worker without having access to the entire graph~\cite{Zhu2023}. They point out that it is attributed to the discrepancy of data distributions caused by graph partitioning algorithms, such as METIS~\cite{karypis1998multilevelk}. It is because such algorithms are designed to minimize the edge cut (i.e., number of edges across partitions), without ensuring the same data distribution for all partitions. They propose randomized partitioning methods to overcome the data discrepancy issue in order to improve the prediction performance of GNNs for link prediction in distributed environments. However, such randomized partitioning methods lead to significant information loss as they highly destroy the underlying graph structure and connectivity, making the neighbors of each node fragmented across partitions. In addition, the importance of ``negative sampling" is overlooked, although it is a key operation in training GNNs for link prediction.

In this work, we first revisit the performance degradation problem in distributed GNN training for link prediction. We unveil that its main root causes rather stem from the information loss caused by graph partitioning algorithms, and how to draw `negative samples,' which are randomly chosen source-destination pairs in the graph that are not directly connected via edges. Specifically, we empirically demonstrate that the performance degradation problem can be resolved by allowing each worker to have access to the entire graph for obtaining the full-neighbor information of each node and drawing negative samples correctly. It is, however, too costly to allow each worker to access the entire graph, clearly limiting the benefits of distributed GNN training. Thus, it becomes crucial to minimize the communication overhead of transferring graph data (including graph structure and node features) while maintaining high link prediction accuracy.

We next develop \n{}, a novel distributed GNN training framework for link prediction, which aims to strike a balance between the communication overhead of transferring graph data and link prediction accuracy. To this end, \n{} uses an efficient graph sparsification algorithm, which is based on an approximation of effective resistance, to remove unimportant edges in each partitioned subgraph. The sparsed partitioned subgraphs are then used by each worker, instead of the entire graph, for drawing negative samples. In addition, \n{} allows each worker to maintain the full-neighbors of each node in its own partitioned subgraph (while the other subgraphs are sparsed ones for negative sampling) to eliminate the information loss caused by graph partitioning. We conduct extensive experiments and show that \n{} reduces the communication overhead by up to 80\% while mostly preserving link prediction accuracy. We summarize our contributions as follows:

\begin{itemize}[itemsep=1pt,leftmargin=1.1em,topsep=1pt]
\item We demystify the performance of distributed training of GNNs for link prediction by uncovering the main sources of the performance drop problem, which are the information loss caused by graph partitioning and how to generate negative samples under distributed settings.

\item We propose \n{} that effectively leverages graph sparsification to reduce the communication overhead while achieving high prediction accuracy.

\item We evaluate the efficiency and effectiveness of \n{} on nine public real-world datasets and demonstrate its superior performance over state-of-the-art distributed GNN training methods for link prediction.

\end{itemize}
\section{Preliminaries}

\subsection{Link Prediction}

\begin{figure}[t]
    \captionsetup[subfloat]{captionskip=1pt}
    \vspace{0mm}
    \centering
    \subfloat[]{%
        \includegraphics[width=0.42\linewidth, trim=0cm 0cm 0cm 0cm, clip]{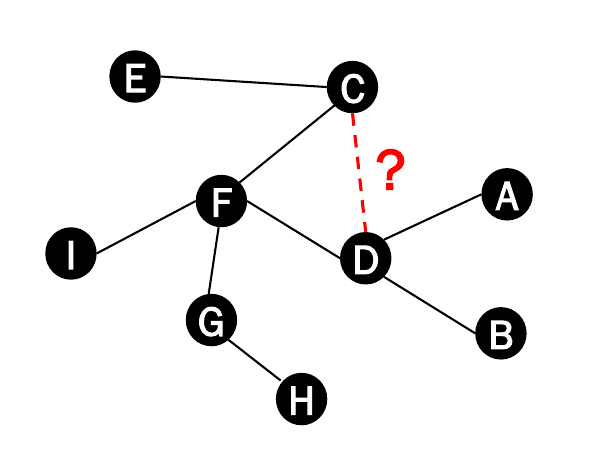}
    }
    \hspace{-1mm}
    \subfloat[]{%
        \includegraphics[width=0.42\linewidth, trim=0cm 0cm 0cm 0cm, clip]{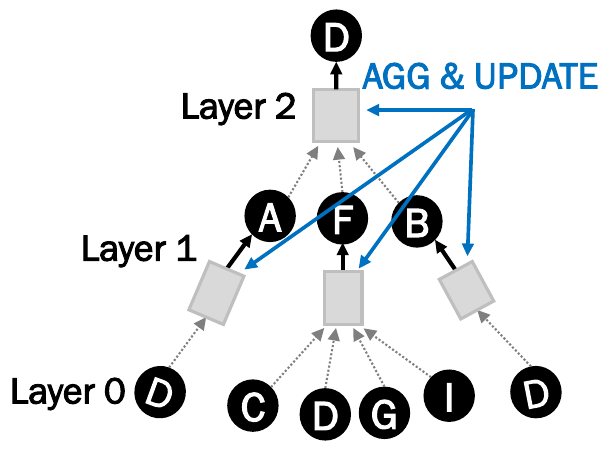}
    }
    \vspace{-0.5mm}
    \caption{Illustration of (a) link prediction and (b) the computational graph of node $D$.}
    \label{fig:linkpred}
    \vspace{0mm}
\end{figure}

Consider a graph $\mathcal{G} \!=\! (\mathcal{V}, \mathcal{E}, \mathbf{X})$, where $\mathcal{V}$ is the node set, $\mathcal{E} \subseteq \mathcal{V} \times \mathcal{V}$ is the edge set, and $\mathbf{X} = [\bm{x}_1, \bm{x}_2, \ldots, \bm{x}_{|\mathcal{V}|}] \in \mathbb{R}^{f \times |\mathcal{V}|}$ is the matrix of initial node features, with $\bm{x}_i \in \mathbb{R}^f$ being the initial feature vector of node $i$. Link prediction~\cite{wu2022graph} is to learn from $\mathcal{G}$ to predict the existence of a missing or potential link between a pair of nodes $u$ and $v$, as illustrated in Figure~\ref{fig:linkpred}(a). The link prediction has been extensively studied in the literature and widely used for social networks, knowledge graphs, and recommender systems~\cite{ying2018graph,schlichtkrull2018modeling}.

There have been several algorithms and methods to solve the problem of link prediction. Classical approaches mostly rely on heuristics that define similarity scores to evaluate the likelihood of the presence of a link. Such examples include common neighbors, Jaccard index, and preferential attachment~\cite{wu2022graph}. Network embedding algorithms, such as DeepWalk~\cite{perozzi2014deepwalk} and node2vec~\cite{grover2016node2vec}, have also been used for link prediction. They first learn node embeddings, e.g., via random walks, and then aggregate pairwise node embeddings as link embeddings for link prediction. 

\subsection{Graph Neural Networks for Link Prediction}\label{se:gnn-lp}

Link prediction has recently been considered as an important application of GNNs due to their ability in learning from graph-structured data. The principle of GNNs is the neighborhood aggregation mechanism~\cite{hamilton2020graph}. Specifically, at each iteration, the embedding of each node is updated by aggregating the embeddings of its neighbors and its previous embedding. Figure~\ref{fig:linkpred}(b) shows the computational graph for node $D$ with a two-layer GNN to illustrate the process of the neighborhood aggregation to update the embedding of node $D$ when the two-layer GNN is applied to the graph in Figure~\ref{fig:linkpred}(a). 

Formally, for a $K$-layer GNN, the embedding (vector) of node $v$ at the $k$-th layer, say, $\bm{h}^{(k)}_{v}$, is updated by the following iterative operation:
\begin{equation}
\setlength{\abovedisplayskip}{5pt}
\setlength{\belowdisplayskip}{5pt}
\bm{h}^{(k)}_{v} \!=\! \text{UPDATE}^{(k)}\Big(\bm{h}^{(k\!-\!1)}_{v}, ~\text{AGG}^{(k)}\big(\bm{h}^{(k\!-\!1)}_{u}, \forall u  \!\in\! N(v)\big) \Big),
\label{eqn:gnn_train}
\end{equation}
where $\text{AGG}^{(k)}(\cdot)$ and $\text{UPDATE}^{(k)}(\cdot)$ are the aggregation and update functions at the $k$-th layer, respectively, and $N(v)$ is the set of $v$'s neighbors. We set $\bm{h}^{(0)}_{v} := \bm{x}_{v}$, which is the initial feature vector of node $v$, i.e., the $v$-th column of $\mathbf{X}$.

When it comes to the link prediction problem, after obtaining the node embeddings from the $K$-layer GNN, pairwise node embeddings are aggregated and go through an edge predictor, e.g., dot product or MLP. Its output is used for link prediction. Specifically, the edge score for a pair of nodes $u$ and $v$, denoted as $s_{(u,v)}$, is computed as
\begin{equation}
\setlength{\abovedisplayskip}{5pt}
\setlength{\belowdisplayskip}{5pt}
s_{(u,v)} = \text{EdgePredictor} \big(\bm{h}_{u}^{(K)}, \bm{h}_{v}^{(K)} \big),
\label{eqn:gnn_link}
\end{equation}
which is then used to evaluate the likelihood of the presence of an edge (link) between $u$ and $v$.

One important and necessary operation in training a GNN for link prediction is ``negative sampling'', which is to sample negative samples for model training and testing~\cite{yang2024does}. Here, negative samples are the data samples that do not belong to the target class. In other words, for the link prediction problem, negative samples are node pairs that are not connected via edges in the graph. On the contrary, positive samples are the data samples of the target class, i.e., node pairs connected by edges in the graph for the link prediction problem. Positive and negative samples are labeled as 1 and 0, respectively. We can define the sets of positive samples $\mathcal{P}$ and negative samples $\mathcal{N}$ in a graph $\mathcal{G}$ as follows:
\vspace{-1pt}
\begin{align*}
\mathcal{P} &= \{(u, v) \in \mathcal{V} \times \mathcal{V} \mid (u, v) \in \mathcal{E}\}, \\
\mathcal{N} &= \{(u, v) \in \mathcal{V} \times \mathcal{V} \mid (u, v) \notin \mathcal{E}, \, u \neq v\}.
\end{align*}
The sets $\mathcal{P}$ and $\mathcal{N}$ are disjoint, i.e., $ \mathcal{P} \cap \mathcal{N} = \emptyset$ and together represent all possible node pairs, i.e., $ \mathcal{P} \cup \mathcal{N} = \{(u, v) \in \mathcal{V} \times \mathcal{V} \mid u \neq v\} $. 

\begin{figure}[t]
    \vspace{-1mm}
    \centering
    \includegraphics[width=\linewidth, trim=0mm 0mm 0mm 0mm, clip]{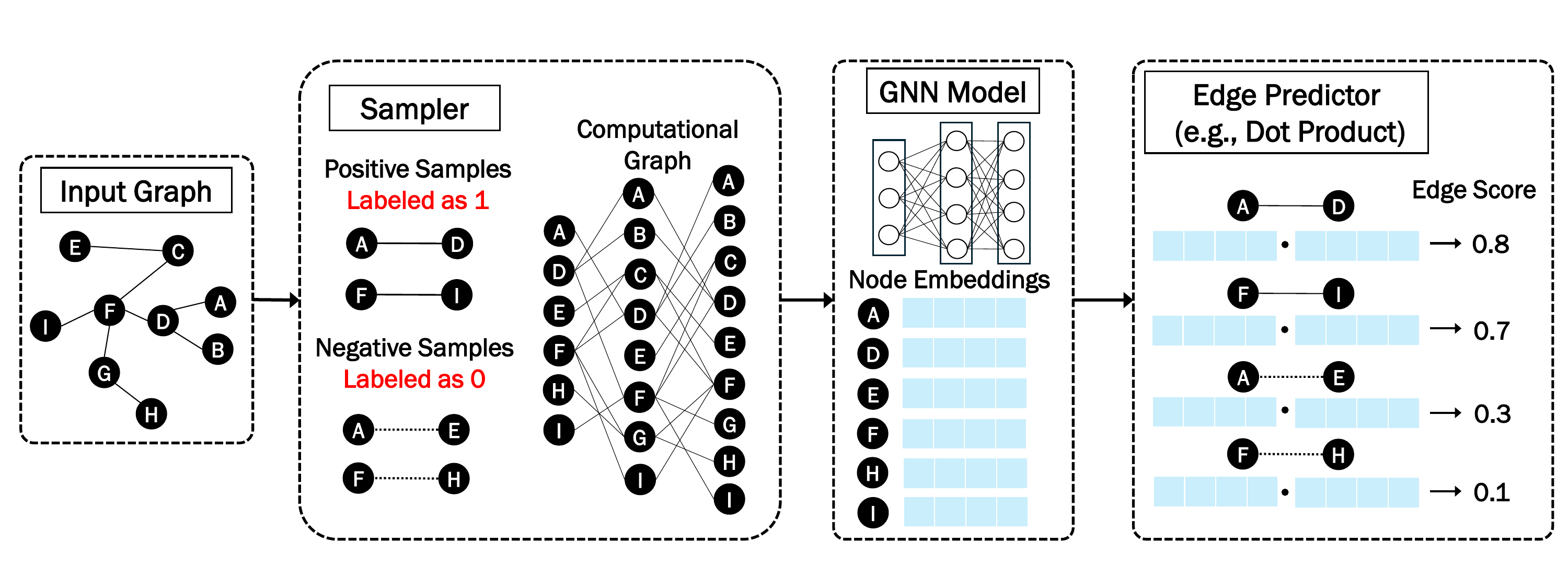}
    \vspace{-8mm}
    \caption{Illustration of GNNs for link prediction.}
    \vspace{0mm}
    \label{fig:GNNlinkpred}
\end{figure}

In practice, $|\mathcal{N}|$ is often much bigger than $|\mathcal{P}|$. Thus, instead of using all node pairs in $\mathcal{N}$ for GNN model training and testing, \emph{negative sampling} is used to sample a subset of node pairs from $\mathcal{N}$ uniformly at random to make a \emph{balanced} dataset. Specifically, there are two common ways for drawing negative samples~\cite{yang2020understanding}. One is the `global' uniform approach, which randomly chooses negative source-destination pairs (node pairs from $\mathcal{N}$) according to a uniform distribution. It is often used for testing purposes. Another one is the `per-source' uniform approach, which randomly chooses negative `destination' nodes for each `source' node according to a uniform distribution. Note that the negative destination nodes are the ones that do not have edges with the source node. It is often used for model training.

Figure~\ref{fig:GNNlinkpred} illustrates the entire pipeline of using a GNN for solving the link prediction problem. For model training, the sampler generates a batch of positive and negative samples and constructs a computational graph based on the selected samples as well as the structure of the input graph. Node embeddings are then computed based on the computational graph and the GNN model according to Eq.~\eqref{eqn:gnn_train}. The edge predictor takes the pairwise node embeddings and computes the edge score for each sample (each pair of nodes), as in Eq.~\eqref{eqn:gnn_link}. The loss between the output edge scores and ground-truth labels (1 for positive edges and 0 for negative edges) is computed based on a predefined loss function, such as binary cross-entropy. The gradients are then obtained based on the loss and are backpropagated through the model (i.e., GNN with the edge predictor). Finally, model weights/parameters are updated based on the gradients. This process is repeated iteratively until convergence or reaching a predefined maximum number of training epochs. Once trained, the GNN model along with the edge predictor can be used to predict the existence of an edge between two nodes. The higher the output edge score, the more likely there is an edge between the node pair.

\subsection{Distributed GNN Training}

Several distributed GNN training systems and frameworks~\cite{FeyLenssen2019,zheng2020distdgl,zheng2022distributed,md2021distgnn,liu2023bgl,wan2023adaptive} have been developed to accelerate training of GNNs and enhance their scalability on large graphs, although they are mostly designed for node classification. The general pipeline of distributed GNN training for link prediction works as follows: A master server is used to partition the input graph (and its associated node features) by a graph partitioning algorithm like METIS~\cite{karypis1998multilevelk}. The resulting subgraphs, along with their node features, are then distributed to workers in the cluster. Each worker processes its allocated subgraph for GNN training. In addition, a global model is initialized and copied to each worker. During model training, the sampler in each worker iteratively generates a batch of positive and negative samples (edges) based on its assigned subgraph and constructs the corresponding computational graph for updating node embeddings using Eq.~\eqref{eqn:gnn_train}. The edge scores are also computed based on pairwise node embeddings, as in Eq.~\eqref{eqn:gnn_link}. The trained local model replicas from all workers are synchronized using gradient or model synchronization methods. Note that workers may need to communicate with each other or the master server to have full access to the entire graph and node features. 

\section{Performance Drop in Link Prediction Using Distributed GNN Training}\label{sec:existing}

We below review state-of-the-art methods to mitigate the problem of a performance drop in link prediction when using distributed GNN training. We then unveil the \emph{true} root causes of the performance drop problem and discuss the technical challenges associated with the problem.

\begin{figure*}[t!]
    \centering
    \subfloat[PSGD-PA]{%
        \includegraphics[width=0.4\linewidth, trim=0cm 0cm 0cm 0cm, clip]{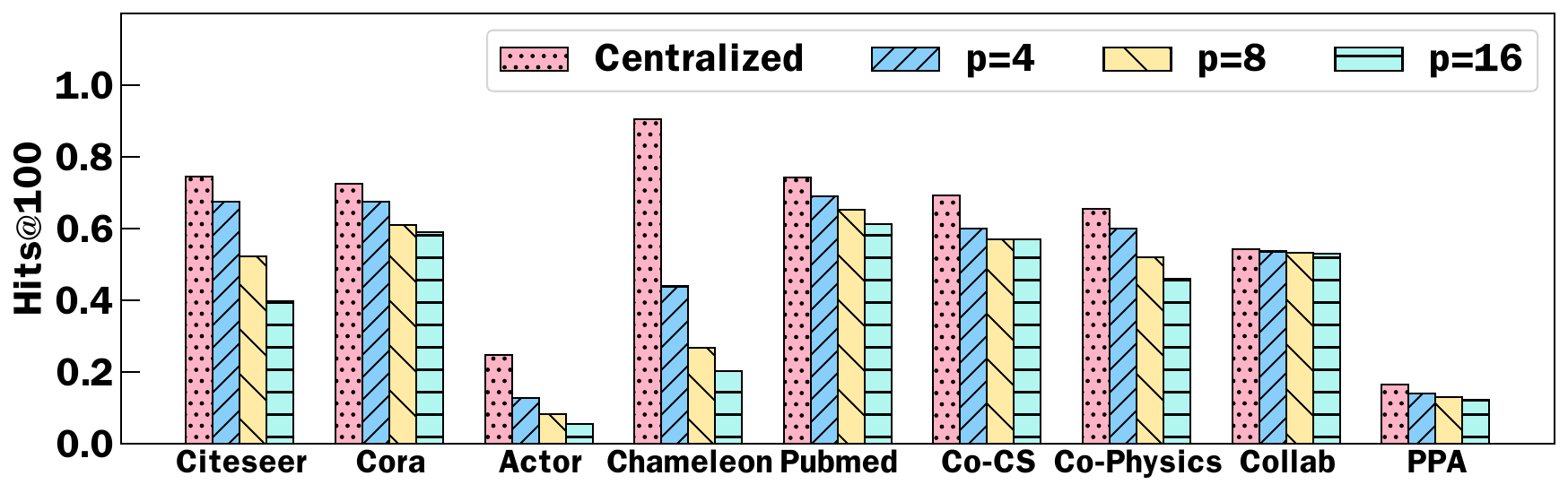}
    }
    \hspace{1mm}
    \subfloat[LLCG]{%
        \includegraphics[width=0.4\linewidth, trim=0cm 0cm 0cm 0cm, clip]{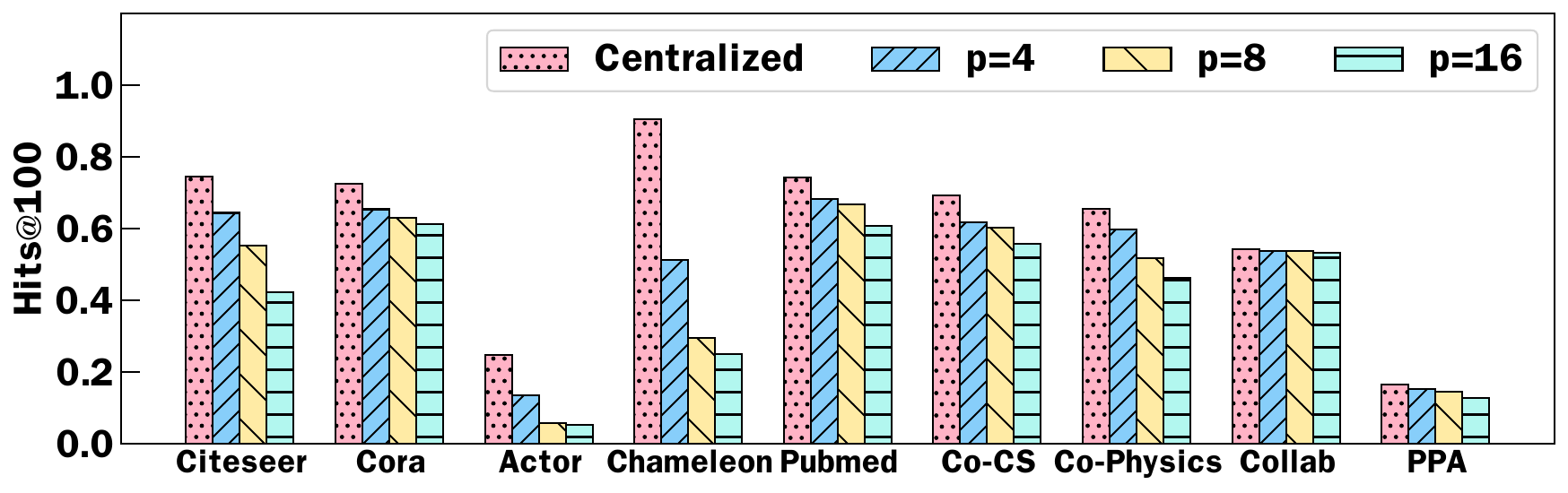}
    }
    \vspace{1mm}
    \subfloat[RandomTMA]{%
        \includegraphics[width=0.4\linewidth, trim=0cm 0cm 0cm 0cm, clip]{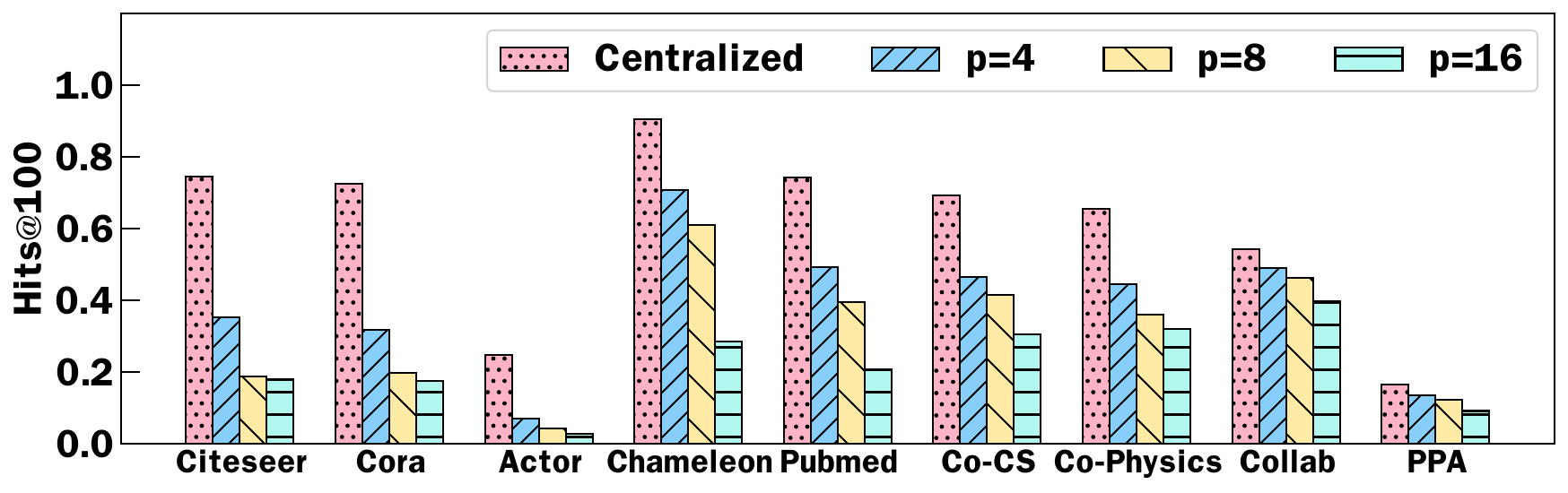}
    }
    \hspace{1mm}
    \subfloat[SuperTMA]{%
        \includegraphics[width=0.4\linewidth, trim=0cm 0cm 0cm 0cm, clip]{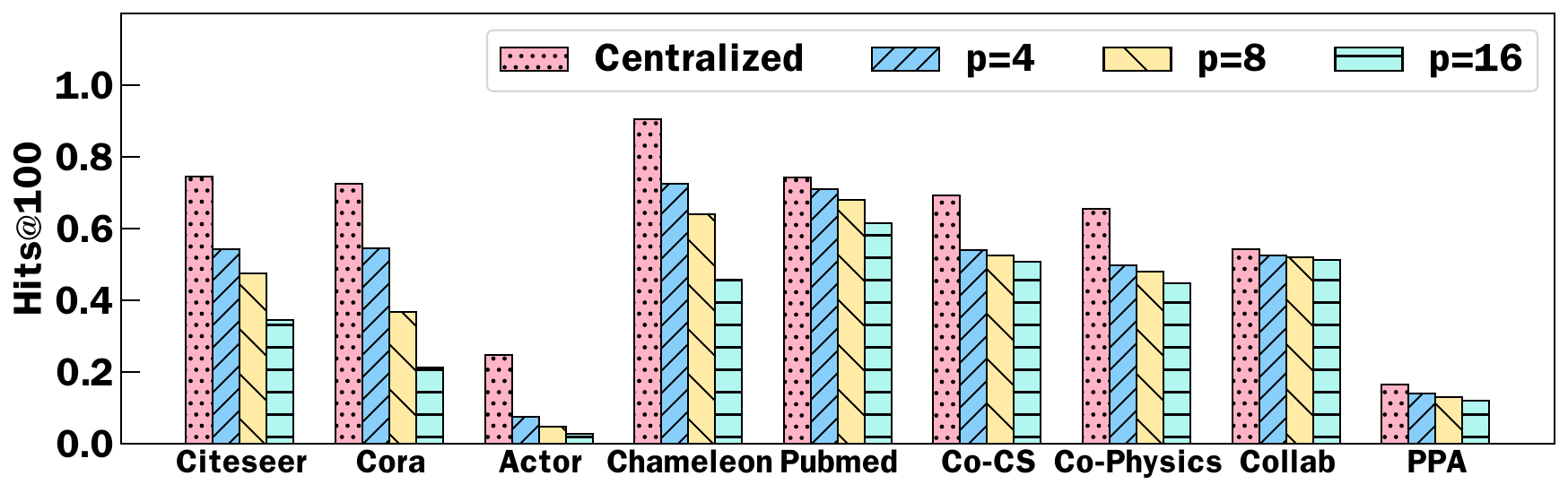}
    }
    \vspace{-0.5mm}
    \caption{Link prediction accuracy of GraphSAGE models trained by the state-of-the-art methods.}
    \label{fig:graphsage}
\vspace{-1mm}
\end{figure*}

\begin{figure*}[ht]
\captionsetup[subfloat]{captionskip=1pt}
\centering
    \subfloat[PSGD-PA+; Accuray]{%
        \includegraphics[width=0.325\linewidth, trim=0cm 0cm 0cm 0cm, clip]{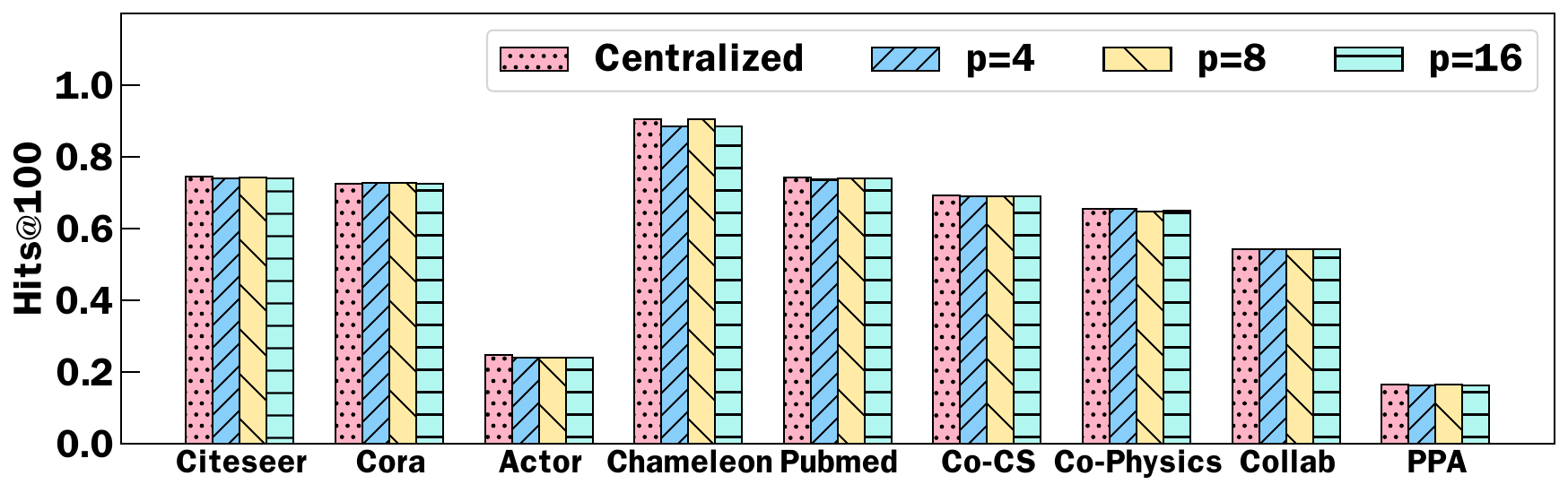}
    }
    \hspace{-1mm}
    \subfloat[RandomTMA+; Accuray]{%
        \includegraphics[width=0.325\linewidth, trim=0cm 0cm 0cm 0cm, clip]{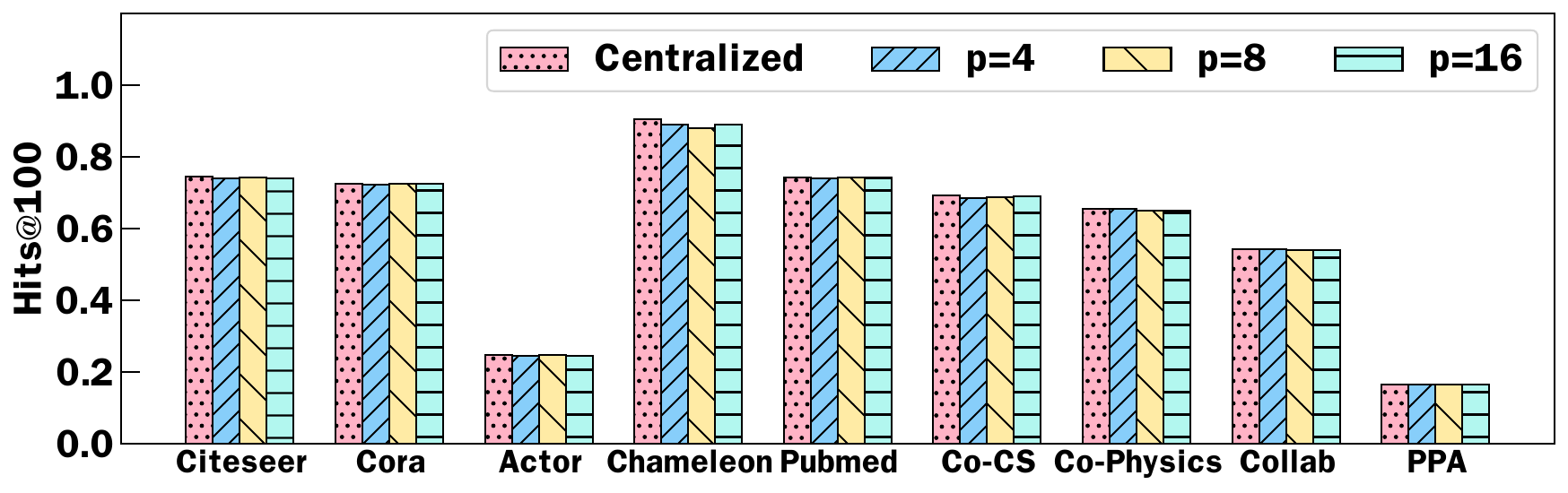}
    }
    \hspace{-1mm}
    \subfloat[SuperTMA+; Accuray]{%
        \includegraphics[width=0.325\linewidth, trim=0cm 0cm 0cm 0cm, clip]{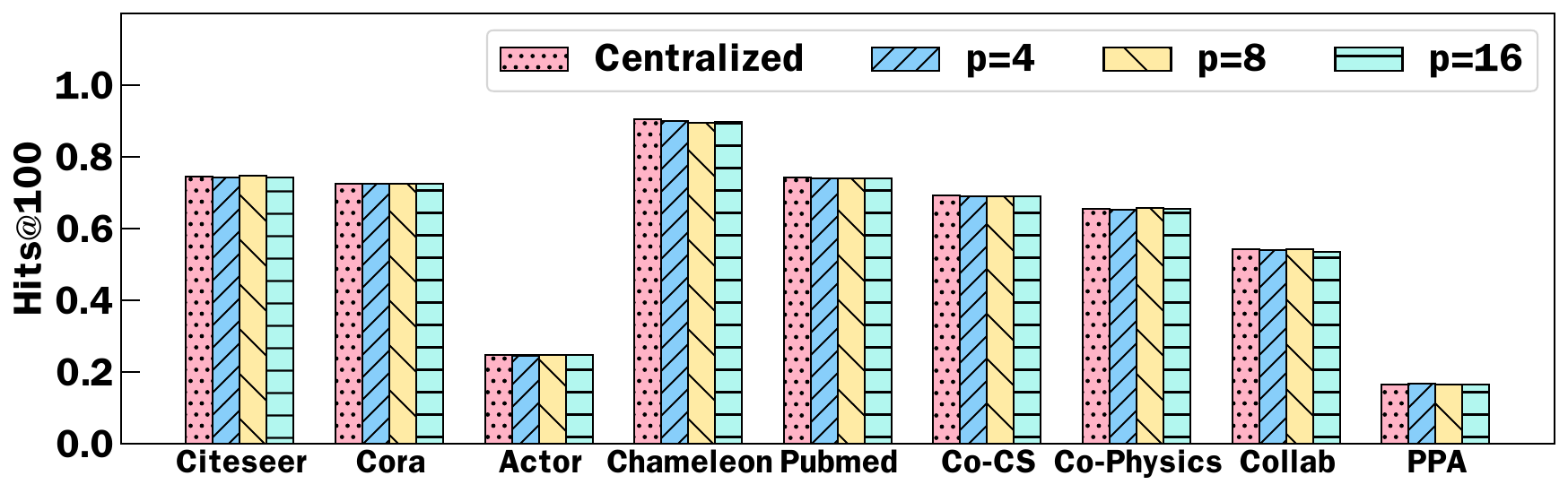}
    }
    \vspace{1mm}
    \subfloat[PSGD-PA+; Communication cost]{%
        \includegraphics[width=0.325\linewidth, trim=0cm 0cm 0cm 0cm, clip]{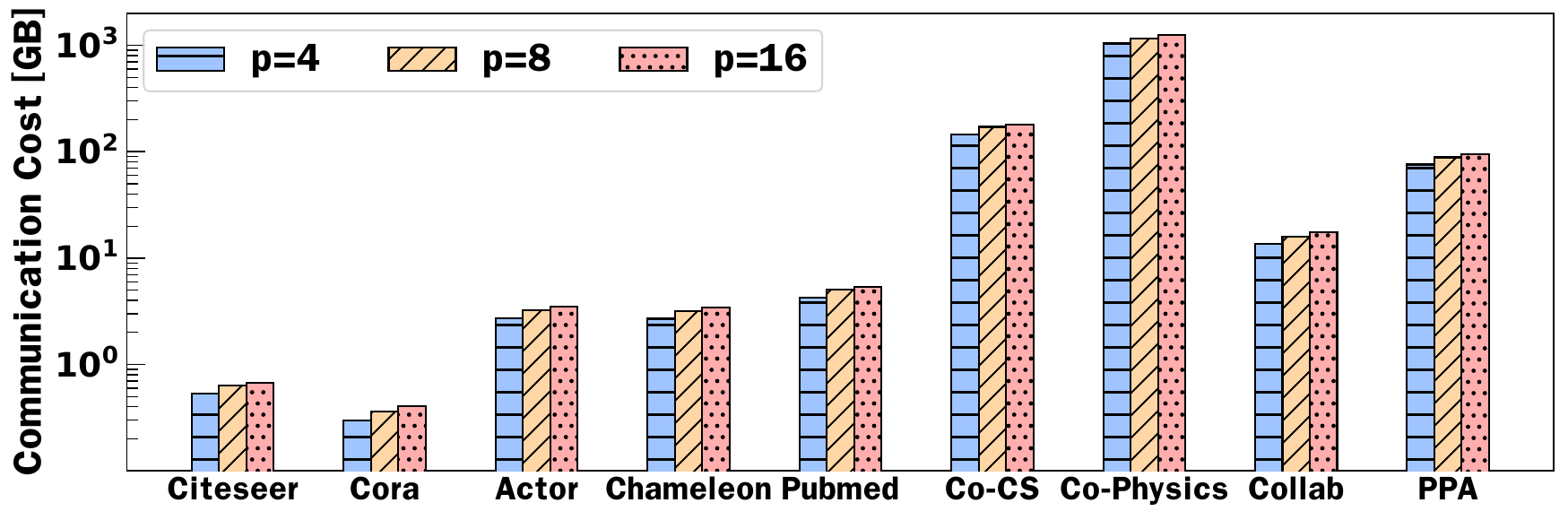}
    }
    \hspace{-1mm}
    \subfloat[RandomTMA+; Communication cost]{%
        \includegraphics[width=0.325\linewidth, trim=0cm 0cm 0cm 0cm, clip]{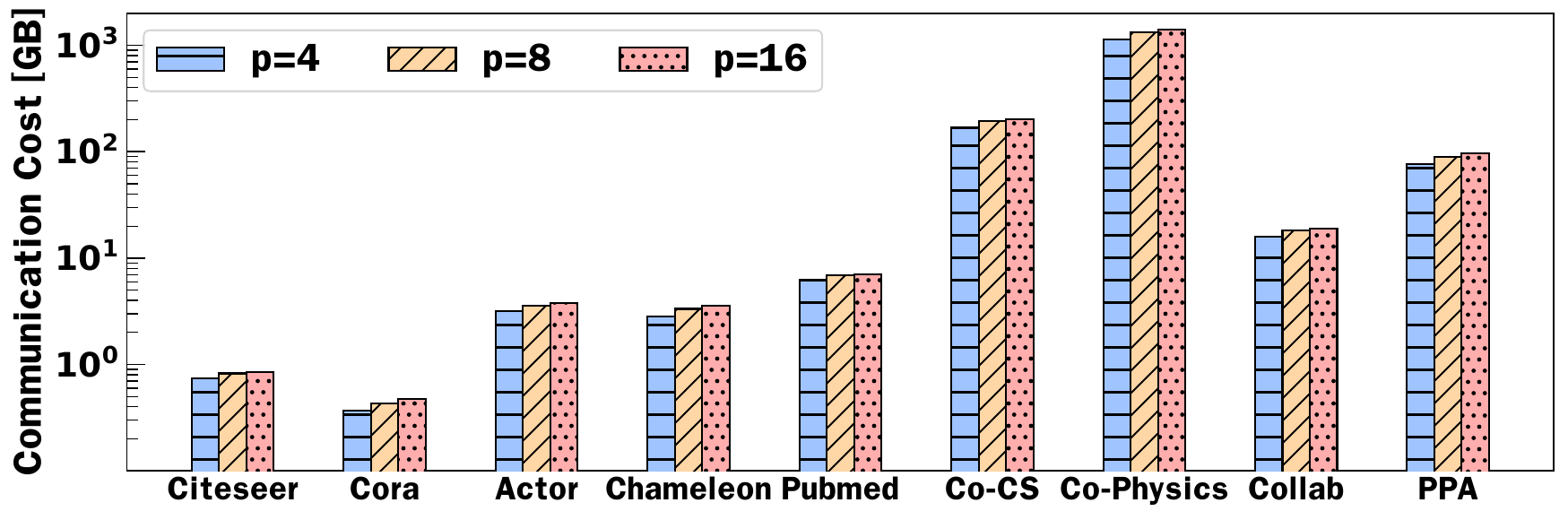}
    }
    \hspace{-1mm}
    \subfloat[SuperTMA+; Communication cost]{%
        \includegraphics[width=0.325\linewidth, trim=0cm 0cm 0cm 0cm, clip]{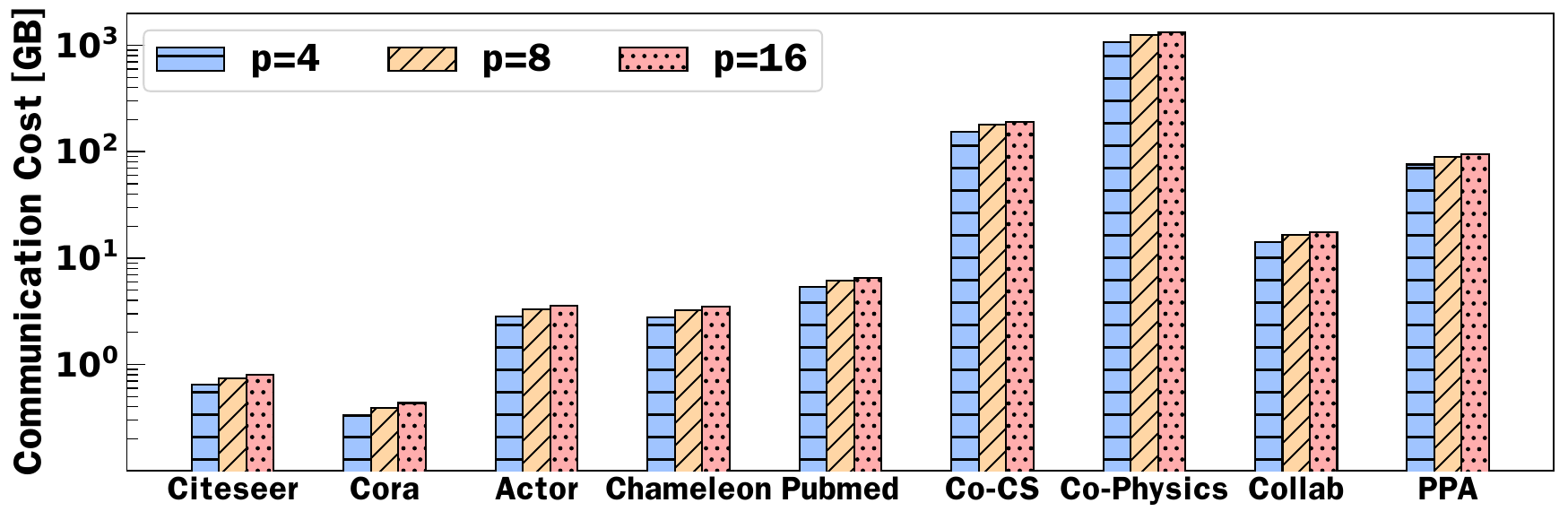}
    }
    \vspace{-0.5mm}
\caption{Accuracy and communication cost of the state-of-the-art methods with the complete data-sharing strategy.}
\label{fig:gn_acc_cost}
\vspace{-5mm}
\end{figure*}

\begin{figure}[t]
    \vspace{0mm}
    \centering
    \includegraphics[width=0.9\linewidth, trim=0mm 0mm 0mm 0mm, clip]{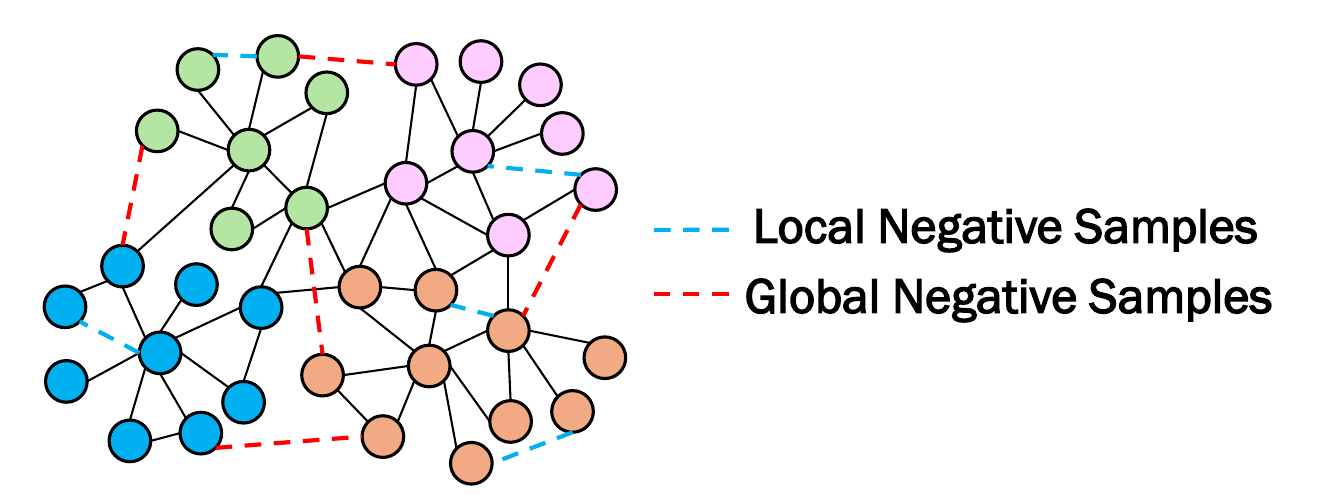}
    \vspace{-3mm}
    \caption{Negative samples.}
    \vspace{-1mm}
    \label{fig:negative}
\end{figure}

\subsection{State-of-the-art Methods}

A recent work~\cite{Zhu2023} points out that the performance of link prediction degrades when using distributed GNN training. They claim that the main reason for such a performance drop is the discrepancy of data distributions across different partitions obtained by a graph partitioning algorithm, such as METIS, rather than the information loss due to ignoring cross-partition edges. To address the data discrepancy, they propose two randomized partitioning methods, namely, RandomTMA and SuperTMA. The former partitions the graph by assigning each node randomly and independently to one of the partitions, and a node-induced subgraph forms each partition. The latter first uses METIS to partition the graph into small mini-clusters. Each mini-cluster then is treated as a super-node and is randomly assigned to a partition.

Due to their randomized nature, both methods eliminate the data discrepancy issue. However, we observe that their performance is still \emph{unsatisfactory}, as shown in Figure~\ref{fig:graphsage}, where we also report the results of PSGD-PA~\cite{ramezanilearn2022} and LLCG~\cite{ramezanilearn2022}, as done in the work~\cite{Zhu2023}. Note that PSGD-PA and LLCG were originally proposed for node classification, but they are modified for link prediction. PSGD-PA partitions the graph using METIS, and each worker trains a GNN model on its assigned subgraph. The models trained by all workers are synchronized periodically via model averaging. In addition, LLCG introduces a global correction step to mitigate the issue of information loss, caused by ignoring the cross-partition edges.\footnote{The global correction step of LLCG requires centralized training on the entire graph in the master server. In that sense, LLCG is not a \emph{pure} distributed training method.} Nonetheless, as shown in Figure~\ref{fig:graphsage}, there is a clear performance degradation when compared with the GNN model trained in a centralized manner.

\subsection{Importance of Negative Samples}

Unlike the node classification using GNNs, in which the embedding of each node is updated based only on the embeddings of its $k$-hop neighbors, the link prediction involves learning from negative samples, where the two endpoint nodes of each negative sample (edge) are often far away from each other. The embeddings of the two endpoints of each negative sample are learned jointly to form an edge embedding for link prediction. In other words, it requires not only the local information, i.e., the $k$-hop neighbors of each node, but also a global view of the graph structure for generating negative samples.

When it comes to distributed training, the entire graph is partitioned into a given number of subgraphs. If each worker only operates on its assigned local subgraph without knowing the entire graph structure, only the local negative samples, i.e., node pairs whose source and destination nodes are from the same partition, can be generated and used for model training, as shown in Figure~\ref{fig:negative}. The other global negative samples, i.e., node pairs whose source and destination nodes are from different partitions, cannot be selected, thereby resulting in a significant information loss. In other words, the `sample space' for each negative sample is only limited to a partitioned subgraph, not the entire graph. This issue leads to a prediction accuracy degradation, as shown in Figure~\ref{fig:graphsage}. We here only report the results of GraphSAGE as a representative example due to the space constraint. We tested different GNN models and observed that they show a similar trend.

A simplest solution to avoid the information loss would be to allow each worker to have access to the entire graph data, including graph structure and node features, by communicating with the master server during model training. To evaluate its performance, we create a shared memory on the master server to share the entire graph data between workers. During model training, instead of only operating on its assigned local subgraph, each worker is allowed to access the shared memory for full-graph information to draw negative samples globally and to obtain the full $k$-hop neighbors (with node features) of each node when they are not locally available.

We apply this complete data-sharing strategy to the above state-of-the-art methods. We observe that they now achieve the same prediction accuracy as that of the GNN model trained in a centralized fashion, but the communication cost of transferring graph data becomes excessively high, as shown in Figure~\ref{fig:gn_acc_cost}. We use PSGD-PA+, RandomTMA+, and SuperTMA+ to denote PSGD-PA, RandomTMA, and SuperTMA when used with the complete data-sharing strategy, respectively, to be distinguished from their vanilla counterparts. Note that we here do not include LLCG due to its correction step that already uses the global information. With the complete data-sharing strategy, the correction step becomes redundant, and it turns out to be the same as PSGD-PA. To measure the communication cost, we collected the total cumulative amount of data transferred from the master server to all workers for one training epoch in gigabyte (GB).

We observe that the total amount of data to be transferred can up to thousands of GB. This is because, for each training iteration, i.e, mini-batch, each worker needs to obtain the full $k$-hop neighbors of the endpoint nodes as well as their features of positive samples if they are not locally available, since the full neighbor information of each node is fragmented by the graph partitioning algorithm. In addition, the endpoint nodes of each negative sample that belongs to different partitions, along with their full $k$-hop neighbors and node features, are also needed to be transferred from the master server to each worker. In other words, the communication cost of the complete data-sharing strategy in distributed GNN training for link prediction is simply too high.

In summary, without sharing the graph data fully, the link prediction accuracy is unsatisfactory due to the significant information loss that occurs from both graph partitioning and incomplete negative samples, although there is no cost for the graph data transfer during training. In contrast, with the complete data-sharing strategy, the accuracy turns out to be identical to the one obtained by centralized training, but the communication overhead becomes too high. Thus, it calls for a cost-efficient distributed training solution to achieve high link prediction accuracy at a reduced communication cost.

\section{\n{}}

We propose \n{}, a novel distributed training framework with graph \textbf{sp}arsification for \textbf{l}ink \textbf{p}rediction via \textbf{G}NNs. We introduce a simple yet effective sparsification approach in \n{} to reduce the communication overhead of graph data transfer during distributed model training while maintaining high link prediction accuracy.

\subsection{Leveraging Graph Sparsification}

Graph sparsification algorithms have been proposed to sparsify a graph $\mathcal{G}$ by removing unimportant or unnecessary edges such that the resulting `sparsed' graph $\mathcal{\Tilde{G}}$ still maintains certain properties of the original graph $\mathcal{G}$~\cite{chen2023demystifying}. Among others, the effective resistance-based graph sparsification algorithm has been widely used in the literature due to its theoretical guarantees~\cite{spielman2008graph, qiu2021lightne,liu2023dspar,liu2022survey}. The effective resistance is a concept derived from electrical network theory and applied to a graph, which is the total resistance experienced between nodes $u$ and $v$ when the graph is viewed as an electrical circuit and each edge is viewed as a resistor.

Consider an undirected, unweighted graph $\mathcal{G}$. Let $\bm{A} \!=\! [A_{u,v}]  \!\in\! \mathbb{R}^{|\mathcal{V}|\times|\mathcal{V}|}$ be the adjacency matrix, where $A_{u,v} \!=\! 1$ if $(u,v) \!\in\! \mathcal{E}$ and $A_{u,v} \!=\! 0$ otherwise. Let $d_u$ be the degree of node $u$, which is given by $d_u \!=\! \sum_{v} A_{u,v}$. We here focus on unweighted graphs as the datasets used for GNN training are mostly unweighted ones~\cite{hu2020open}. Then, we define the graph Laplacian by $\bm{L} \!=\! \bm{D} \!-\! \bm{A}$ and the normalized graph Laplacian by $\bm{L}_{\text{sym}} \!=\! \bm{D}^{-\frac{1}{2}}\bm{L} \bm{D}^{-\frac{1}{2}}$, where $\bm{D} = [D_{u,u}]$ is the diagonal matrix with $D_{u,u} \!=\! d_u$ for all $u$. The effective resistance $r_{(u,v)}$ of an edge $(u,v)$ in $\mathcal{G}$ is defined as
\begin{equation}
\setlength{\abovedisplayskip}{5pt}
\setlength{\belowdisplayskip}{5pt}
r_{(u,v)} = (\bm{e}_u - \bm{e}_v)^T \bm{L}^+ (\bm{e}_u - \bm{e}_v),
\label{eqn:Re}
\end{equation}
where $\bm{L}^+$ is the pseudo-inverse of $\bm{L}$, and $\bm{e}_u$ is the unit vector that has a one at position $u$ and zeros elsewhere.

Spielman and Srivastava~\cite{spielman2008graph} establish that a sparsed graph $\mathcal{\Tilde{G}}$ constructed by selecting a sufficient number of edges with probabilities proportional to their effective resistances is an accurate estimate of $\mathcal{G}$. Letting $\bm{L}$ and $\bm{\Tilde{L}}$ be the Laplacian of $\mathcal{G}$ and $\mathcal{\Tilde{G}}$, respectively, they show the following:

\begin{theorem}\label{theorem:lap}
Sample $n$ edges independently with replacement from $\mathcal{G}$ to construct $\mathcal{\Tilde{G}}$, where each edge $(u,v)$ is chosen with probability $p_{(u,v)}\!\propto\! r_{(u,v)}$ and assigned with a weight $\Tilde{w}_{(u,v)} \!=\! 1/(n p_{(u,v)})$, summing weights if an edge is chosen more than once. If $n\!=\!\mathcal{O}(|\mathcal{V}|\log|\mathcal{V}|/\epsilon^2)$ for some small constant $\epsilon$, then with high probability, for all $\bm{x} \in \mathbb{R}^{|\mathcal{V}|}$,  
\begin{equation*}
\setlength{\abovedisplayskip}{5pt}
\setlength{\belowdisplayskip}{5pt}
(1-\epsilon)\bm{x}^T\bm{L}\bm{x} \leq \bm{x}^T\bm{\Tilde{L}}\bm{x} \leq (1+\epsilon)\bm{x}^T\bm{L}\bm{x}.
\end{equation*}
\end{theorem}

\begin{figure}[t]
    \captionsetup[subfloat]{captionskip=1pt}
    \centering
    \subfloat[GCN]{%
    \includegraphics[width=0.88\linewidth, trim=0cm 0cm 0cm 0cm, clip]{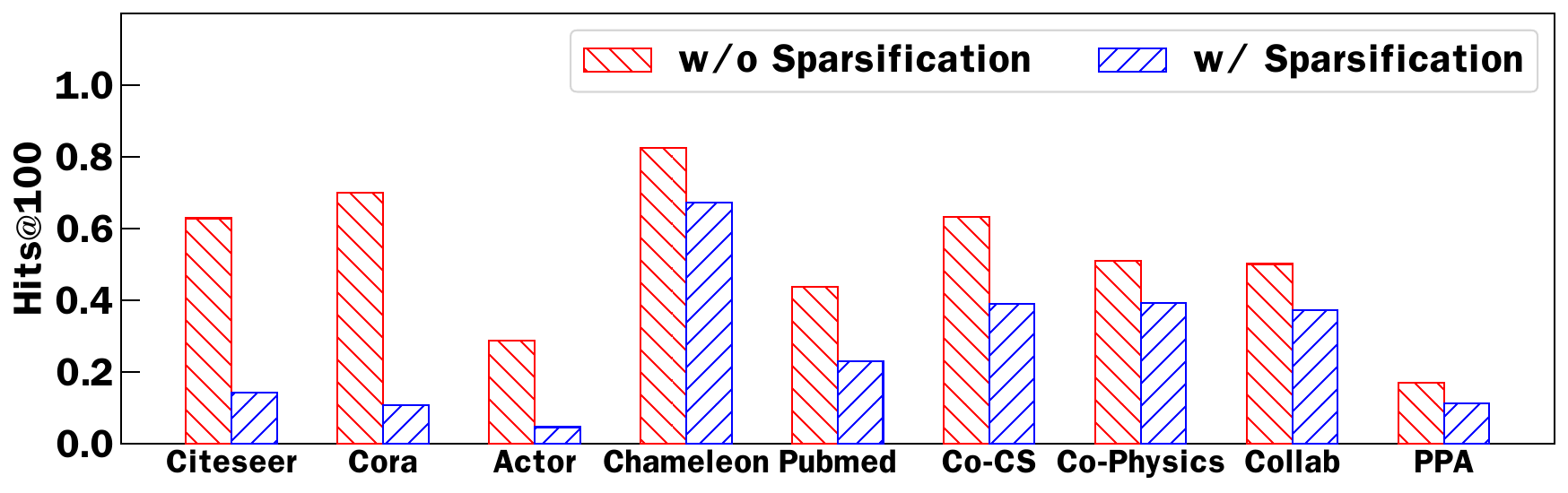}
    }
    \vspace{0mm}
    \subfloat[GraphSAGE]{%
    \includegraphics[width=0.88\linewidth, trim=0cm 0cm 0cm 0cm, clip]{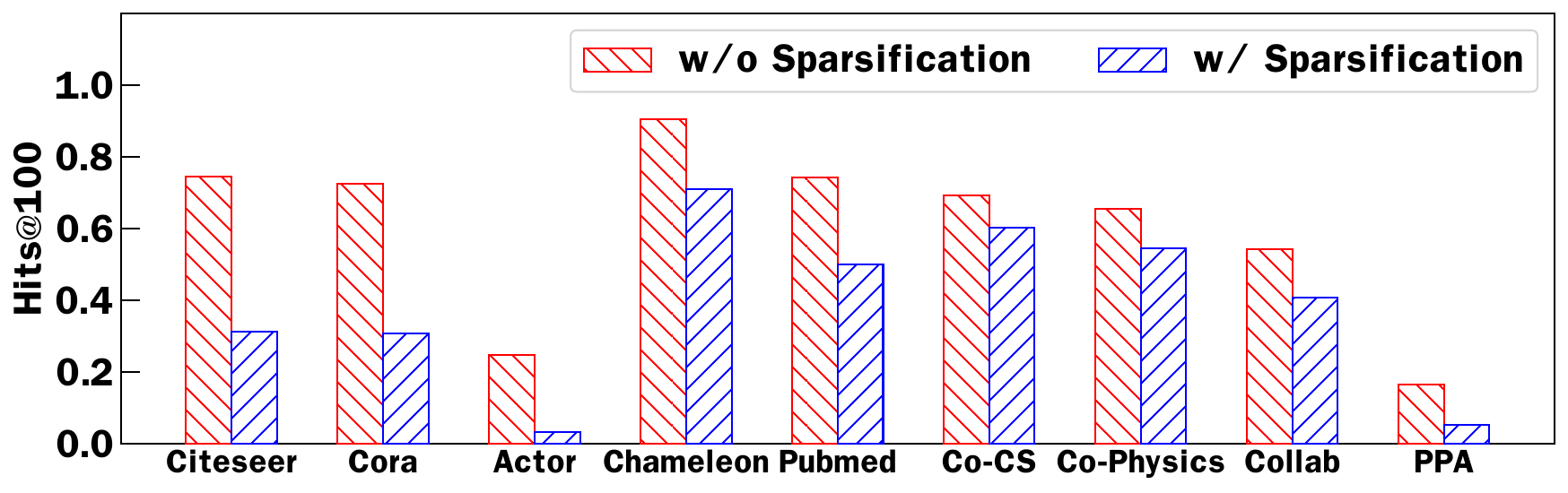}
    }
    \vspace{-0.5mm}
    \caption{Accuracy of GNNs w/ and w/o graph sparsification.}
    \label{fig:DSpar}
    \vspace{-1mm}
\end{figure}

Despite the theoretical guarantees, the direct computation of $r_{(u,v)}$ is expensive since it requires computing the pseudo-inverse $\bm{L}^+$, as can be seen from Eq.~\eqref{eqn:Re}~\cite{chen2023demystifying}. Thus, to effectively approximate $r_{(u,v)}$, we consider the following result:

\begin{theorem}[Corollary 3.3 in~\cite{lovasz1993random}]\label{theorem:er}
For any edge $(u,v) \in \mathcal{E}$, $\frac{1}{2} (\frac{1}{d_u} + \frac{1}{d_v}) \leq r_{(u,v)} \leq \frac{1}{\gamma} (\frac{1}{d_u} + \frac{1}{d_v})$, where $\gamma$ is the second smallest eigenvalue of the normalized graph Laplacian $\bm{L}_{\text{sym}}$.
\end{theorem}
\vspace{0mm}

\begin{figure*}[t!]
    \vspace{0mm}
    \centering
    \includegraphics[width=0.99\linewidth, trim=0mm 0mm 0mm 0mm, clip]{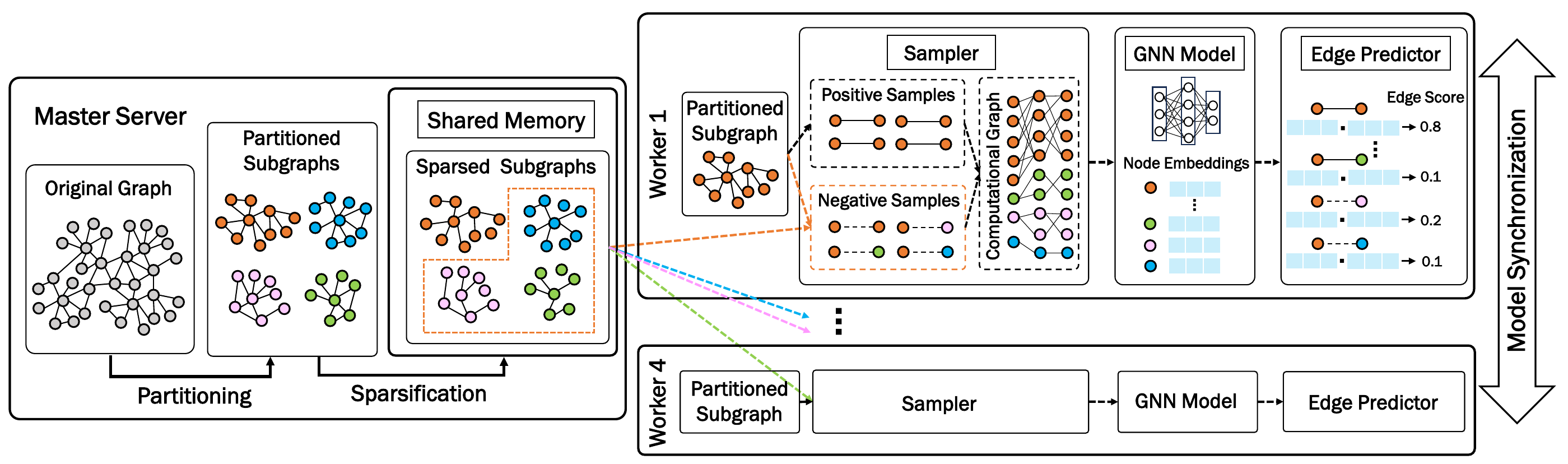}
    \vspace{-2mm}
    \caption{Illustration of \n{}.}
    \vspace{-5mm}
    \label{fig:ours}
\end{figure*}

Theorem~\ref{theorem:er} implies that for each edge $(u,v)$, its effective resistance $r_{(u,v)}$ is bounded by $\frac{1}{d_u} + \frac{1}{d_v}$, which is based only on the degrees of nodes $u$ and $v$. Since node degrees can be easily obtained, this approximation of $r_{(u,v)}$ is straightforward to compute. Thus, from Theorems~\ref{theorem:lap} and~\ref{theorem:er}, the graph sparsification algorithm based on the approximation of effective resistance can be used to sparsify a graph efficiently.

It is worth noting that the effective resistance-based graph sparsification algorithm has been employed as a data preprocessing step for graph mining~\cite{chen2023demystifying,qiu2021lightne} and GNN training for node and graph classification tasks~\cite{liu2023dspar,liu2022survey} in order to improve their scalability. However, they cannot be directly used for link prediction. To see this, we use the effective resistance-based graph sparsification algorithm derived from Theorems~\ref{theorem:lap} and~\ref{theorem:er} to sparsify the graph and train GraphSAGE models on the sparsed graph for link prediction. As shown in Figure~\ref{fig:DSpar}, with sparsification, the link prediction accuracy drops up to 80\% compared to the ones without sparsification. This is because plenty of edges are removed after graph sparsification, which greatly reduces the number of \emph{positive} samples. It leads to a significant information loss and thus performance degradation in link prediction. 

On the contrary, we leverage graph sparsification as an effective means to reduce graph data transfer overhead during distributed GNN training. As illustrated in Figure~\ref{fig:ours}, we sparsify each partitioned subgraph such that all nodes are maintained while unimportant edges are removed from the partitioned subgraph. With its partitioned subgraph, each worker can now have access to the other sparsed subgraphs placed in the shared memory, which are \emph{only} used for generating \emph{negative} samples. Specifically, each worker generates positive samples from its assigned subgraph without sparsification. When it comes to generating negative samples, each worker uses its assigned subgraph (again without sparsification) and the other sparsed subgraphs. This way, each worker can generate not only local negative samples but also global ones at reduced communication cost. Note that the sample space for each negative sample is the \emph{entire} graph, but each (global) negative sample from the outside of the assigned subgraph comes with a \emph{much fewer} number of $k$-hop neighbors since unimportant neighbors of each `negative-sampled' node are removed due to the sparsification.\footnotemark ~Thus, we can effectively reduce the communication cost without sacrificing link prediction accuracy much.

\subsection{Description of \n{}}

\footnotetext{Note that, for each worker, the full-neighbor list of each node is maintained in its corresponding partition (subgraph), although the other subgraphs are sparsed ones, where all nodes still remain the same  (with much fewer edges). Also, as mentioned in Section~\ref{se:gnn-lp}, the per-source uniform approach is used to generate negative samples during model training. Thus, the sample space for selecting the destination node of each negative sample remains identical to that of the original graph, regardless of whether it is from sparsed subgraphs.}

We below explain the detailed operations of \n{}, as depicted in Algorithm~\ref{alg:algorithm}. \n{} first uses METIS to partition the input graph $\mathcal{G}$ into $p$ subgraphs $\mathcal{G}^1, \mathcal{G}^2, \ldots, \mathcal{G}^p$ and copies each partitioned subgraph $\mathcal{G}^i \!=\! (\mathcal{V}^i, \mathcal{E}^i, \mathbf{X}^i)$ to worker $i$, where $\mathbf{X}^i$ is the matrix of initial node features of nodes in $\mathcal{V}^i$ (\textbf{Lines 1--3}). Note that the cross-partition edges are maintained in both partitions. That is, the full-neighbor list of each node is fully preserved in a partitioned subgraph. This strategy can effectively reduce the information loss caused by the graph partitioning algorithm. The effective resistance-based graph sparsification algorithm based on Theorems~\ref{theorem:lap} and~\ref{theorem:er} is then employed in \n{} to sparsify each partitioned subgraph (\textbf{Lines 4--14}).

Specifically, \n{} samples $L^i$ edges with replacement in each partitioned subgraph $\mathcal{G}^i$ to obtain its resulting sparsed graph $\mathcal{\Tilde{G}}^i \!=\! (\mathcal{V}^i, \mathcal{\Tilde{E}}^i, \mathbf{X}^i)$. The sampling probability of an edge $(u,v) \!\in\! \mathcal{E}^i$ is $p_{(u,v)} \!\propto\! \frac{1}{d_u} \!+\! \frac{1}{d_v}$, where $d_u$ and $d_v$ are the degrees of nodes $u$ and $v$, respectively. \n{} assigns the weight ${\Tilde{w}}^{i}_{(u,v)} \!=\! 1 / (L^i p_{(u,v)})$ to each sampled edge and sum the weights up if an edge is chosen more than once. The resulting sparsed subgraphs are placed into the shared memory so that all workers can access them.

\newcommand\mycommfont[1]{\small\ttfamily\textcolor{blue}{#1}}
\SetCommentSty{mycommfont}
\SetKwInOut{KwIn}{Input}
\SetKwInOut{KwOut}{Output}
\newcommand{\ForCustom}[2]{%
    \SetKwFor{ForCustom}{#1}{\textnormal{ do #2}}{}
}

\begin{algorithm}[t!]
\small
\SetAlgoNoEnd
\setstretch{1.1}
\caption{\n{}}\label{alg:algorithm}
\KwIn{$\mathcal{G}\!=\! (\mathcal{V}, \mathcal{E}, \mathbf{X})$, number of partitions $p$, \\ sparsification level $L^i$, number of epochs $E$}
\KwOut{Trained model $\bm{W}$}
\tcc {Graph Partitioning (e.g., METIS)}
{Partition $\mathcal{G}$ into $p$ subgraphs $\mathcal{G}^1, \mathcal{G}^2, \ldots, \mathcal{G}^p$.} \\
{Copy $\mathcal{G}^i$ to worker $i$.} \\
\tcc {Graph Sparsification}
\For{$i = 1$ \textnormal{\textbf{to}} $p$}{
    {$\mathcal{\Tilde{E}}^i \leftarrow \{\}$.}  \\
    \For{$l = 1$ \textnormal{\textbf{to}} $L^i$}{
        {Sample an edge $(u,v) \in \mathcal{E}^i$ with probability $p_{(u,v)}$.} \\
        \If {$(u,v) \notin \mathcal{\Tilde{E}}^i$}
        {$\mathcal{\Tilde{E}}^i \leftarrow \mathcal{\Tilde{E}}^i \cup \{(u,v)\}$, ${\Tilde{w}}^{i}_{(u,v)} \leftarrow \frac{1}{L^i p_{(u,v)}}$.}
        \Else
        {${\Tilde{w}}^{i}_{(u,v)} \leftarrow {\Tilde{w}}^{i}_{(u,v)} + \frac{1}{L^i p_{(u,v)}}$.}
        }
    {$\mathcal{\Tilde{G}}^i  \leftarrow (\mathcal{V}^i, \mathcal{\Tilde{E}}^i, \mathbf{X}^i) $.}}
{Place $\mathcal{\Tilde{G}}^1, \mathcal{\Tilde{G}}^2, \ldots, \mathcal{\Tilde{G}}^p$ into the shared memory.} \\
\tcc {Distributed Model Training}
{Initialize model weights $\bm{W}$ and copy them to each worker.} \\
\For{\textnormal{epoch} $=1$ \textnormal{\textbf{to}} $E$}{
    \For{\textnormal{each batch}} {
        \SetKwFor{ForParallel}{for $i = 1$ \textnormal{\textbf{to}} $p$}{do in parallel}{}
        \ForParallel{}{
        {Generate positive samples $\mathcal{P}^i$ based on $\mathcal{G}^i$.} \\
        {Generate negative samples $\mathcal{N}^i$ based on $\mathcal{G}^i$ and $\mathcal{\Tilde{G}}^j, j = 1, \ldots, p~ (j \neq i)$.} \\
        \For{$k=1$ \textnormal{\textbf{to}} $K$}{
        {Compute $\bm{h}^{(k)}_{v}$ using Eq.~(\ref{eqn:gnn_train}).}}
        \For{\textnormal{each edge} $(u,v) \in \mathcal{P}^i \cup \mathcal{N}^i$}{
        {$y_{(u,v)} \gets \begin{cases}
        \text{1} & \text{if } (u,v) \in \mathcal{P}^i ,\\
        \text{0} & \text{if } (u,v) \in \mathcal{N}^i .\\
        \end{cases}$} \\
        {$s_{(u,v)} \leftarrow \text{EdgePredictor} \Big(\bm{h}_{u}^{(K)}, \bm{h}_{v}^{(K)} \Big)$.}}
        {Compute \!$\mathcal{L}^i$ for all pairs of $y_{(u,v)}$ and $s_{(u,v)}.\!$} \\
        {Compute the gradient $\nabla \mathcal{L}^i(\bm{W})$.}}
        {Synchronize $\nabla \mathcal{L}(\bm{W}) \leftarrow \frac{1}{p}\sum^p_{i=1} \nabla \mathcal{L}^i(\bm{W})$.} \\
        {Update $\bm{W} \leftarrow \bm{W} - \eta \nabla \mathcal{L}(\bm{W})$.}}}
\end{algorithm}

The distributed model training process is shown in \textbf{Lines 15--30}. A GNN model with weights $\bm{W}$ is initialized and copied to each worker. Each worker $i$ operates on its assigned subgraph $\mathcal{G}^i$. For each mini-batch, a sampler (in each worker $i$) generates a set of positive samples, say, $\mathcal{P}^i$, based on $\mathcal{G}^i$ and a set of negative samples, say, $\mathcal{N}^i$, based on $\mathcal{G}^i$ and $\mathcal{\Tilde{G}}^j$ for $j \!=\! 1, \ldots, p ~(j \!\neq\! i)$. Positive and negative samples are labeled as 1 and 0, respectively. The computational graph of each endpoint node of each sample in the current batch is constructed. The embedding of each node in the current batch is computed based on the computational graph and the GNN model, as in Eq.~(\ref{eqn:gnn_train}). For each (positive or negative) sampled edge, the edge score is computed using Eq.~(\ref{eqn:gnn_link}).

Next, the total loss $\mathcal{L}^i$ of a batch for each worker $i$ is computed as $\mathcal{L}^i\!=\!\frac{1}{N} \sum_{(u,v) \in \mathcal{P}^i \cup \mathcal{N}^i}\ell \big(s_{(u,v)},y_{(u,v)}\big)$, where $\ell$ is a loss function, e.g., cross-entropy loss, and $N\!=\!|\mathcal{P}^i|\!+\!|\mathcal{N}^i|$ is the batch size. From $\mathcal{L}^i$, the gradient $\nabla \mathcal{L}^i(\bm{W})$ is computed based on the backward propagation algorithm. The gradient $\nabla \mathcal{L}(\bm{W})$ is then synchronized by taking the average of all gradients $\nabla \mathcal{L}^i(\bm{W})$ for $i=1, 2, \ldots, p$. The model weights are finally updated as $\bm{W} \!=\! \bm{W} - \eta \nabla \mathcal{L}(\bm{W})$, where $\eta$ is the learning rate. Note that model averaging~\cite{li2019convergence} can also be used here instead of gradient averaging for model synchronization. This distributed training process is repeated until convergence or for a predefined number of epochs. 
\section{Evaluation}\label{sec:eval}

In this section, we extensively evaluate the performance of \n{} on nine public datasets to demonstrate its effectiveness and efficiency for link prediction.

\subsection{Experimental Setup}

\noindent \textbf{Implementation.} \n{}\footnote{The code is available at \url{https://github.com/xhuang2016/SpLPG}.} is built on top of PyTorch~\cite{NEURIPS2019_9015} and DGL~\cite{wang2019dgl}. We modify the graph storage, graph partitioning, and graph sampler modules provided by DGL for our usages to store the graph with feature data, partition the graph, and sample the subgraphs for the generation of mini-batches, respectively. We implement the graph sparsification algorithm from scratch. We implement the shared memory using PyTorch \texttt{shared\_memory} primitive. We also implement the distributed training module using PyTorch \texttt{DistributedDataParallel} module and \texttt{all\_reduce} primitive for parallel training and model synchronization. \n{} is currently implemented based on the single-machine multi-GPU scenario but can be easily extended to the multi-machine multi-GPU scenario.

\begin{table}[t!]
\renewcommand{\arraystretch}{1.1}
\vspace{-1.5mm}
\caption{Dataset statistics}
\vspace{-1.5mm}
\label{table:dataset}
\centering
\footnotesize
\begin{adjustbox}{width=0.8\columnwidth,center}
    \begin{tabular}{|c|c|c|c|c|}
    \hline
    Dataset & \# Nodes & \# Edges & \# Features \\
    \hline
    \hline
    Citeseer & 3,327 & 9,228 & 3,703 \\
    \hline
    Cora & 2,708 & 10,556 & 1,433  \\
    \hline
    Actor & 7,600 & 53,411 & 932  \\
    \hline
    Chameleon & 2,227 & 62,792 & 2,325  \\
    \hline
    Pubmed & 19,717 & 88,651 & 500  \\
    \hline
    Co-CS & 18,333 & 163,788 & 6,805 \\
    \hline
    Co-Physics & 34,493 & 495,924 & 8,415 \\
    \hline
    Collab & 235,868 & 1,285,465 & 128  \\
    \hline
    PPA & 576,289 & 30,326,273 & 58  \\
    \hline
    \end{tabular}
\end{adjustbox}
\vspace{0mm}
\end{table}

We use PyTorch and DGL to implement the baselines, i.e., PSGD-PA, RandomTMA, and SuperTMA, by following the code provided in their corresponding papers. Note that, while the mainstream frameworks DistDGL~\cite{zheng2020distdgl,zheng2022distributed} and PyG~\cite{FeyLenssen2019} support distributed GNN training for link prediction, it is unclear how they address the information loss caused by graph partitioning and negative sampling since their complete official examples are not publicly available. To ensure a correct and fair comparison, we exclude them in this work.

\begin{figure*}[ht]
    \captionsetup[subfloat]{captionskip=1pt}
    \centering
    \subfloat[\n{} vs. PSGD-PA+]{%
    \includegraphics[width=0.325\linewidth, trim=0cm 0cm 0cm 0cm, clip]{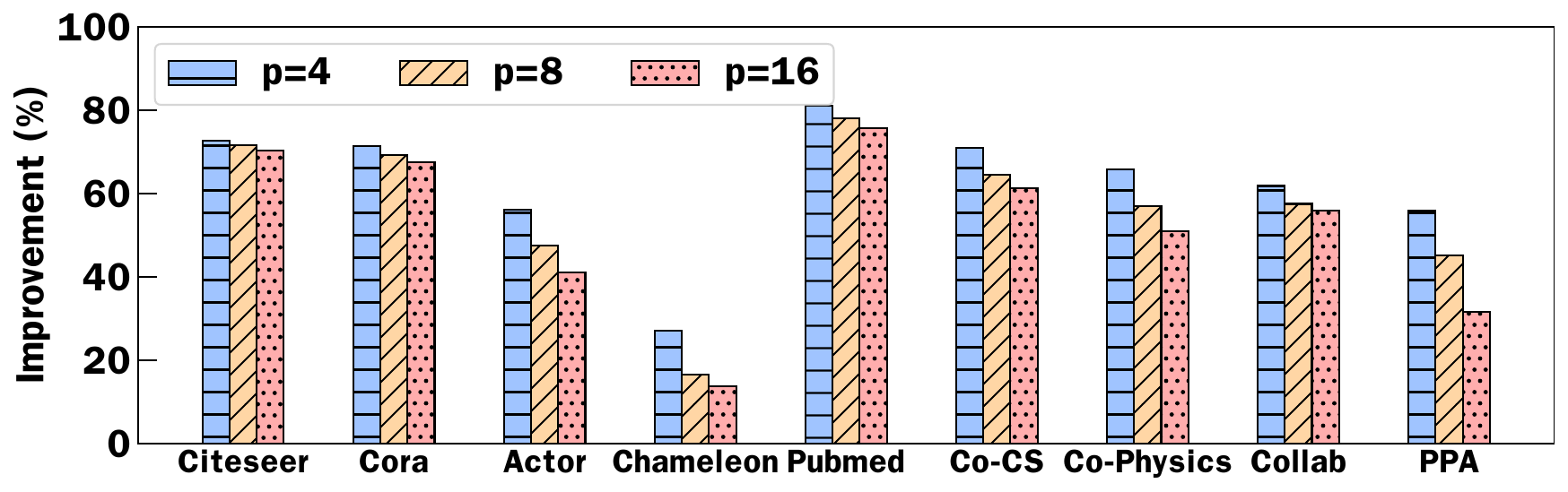}
    }
    \hspace{-2mm}
    \subfloat[\n{} vs. RandomTMA+]{%
    \includegraphics[width=0.325\linewidth, trim=0cm 0cm 0cm 0cm, clip]{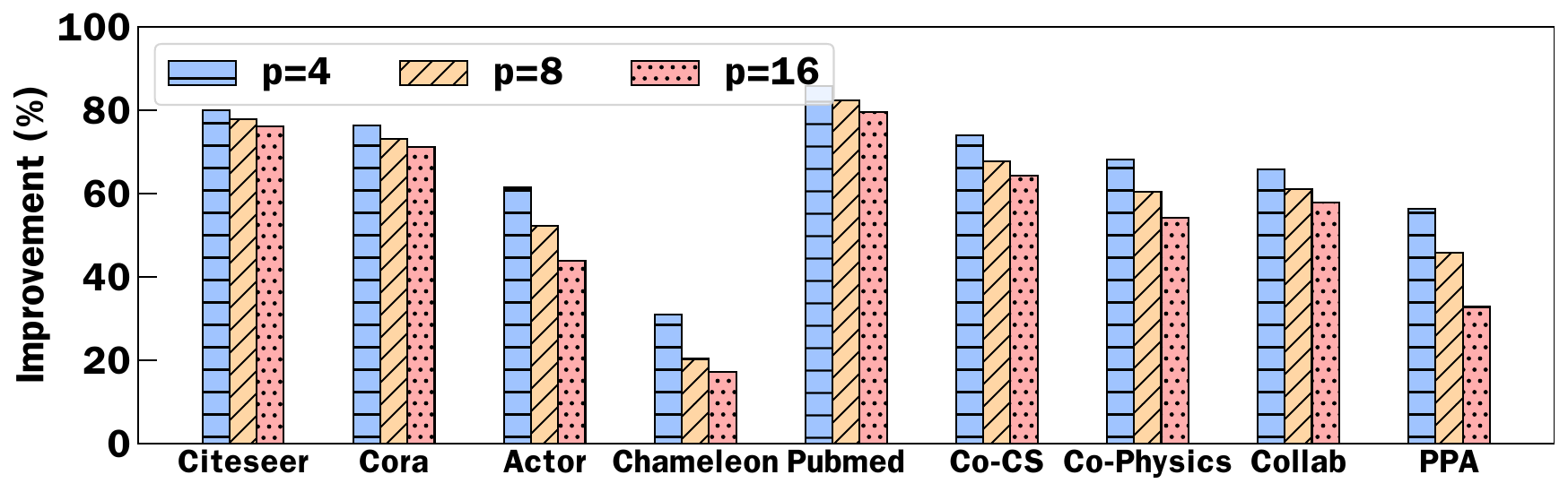}
    }
    \hspace{-2mm}
    \subfloat[\n{} vs. SuperTMA+]{%
    \includegraphics[width=0.325\linewidth, trim=0cm 0cm 0cm 0cm, clip]{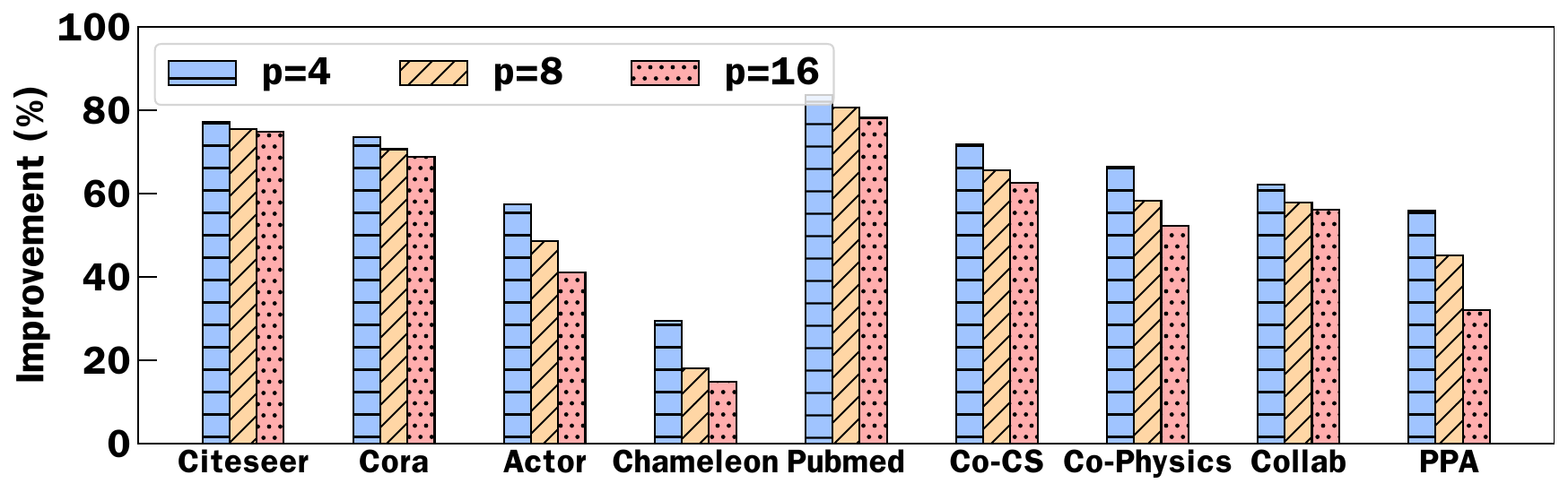}
    }
    \vspace{0mm}
    \subfloat[\n{} vs. PSGD-PA+]{%
    \includegraphics[width=0.32\linewidth, trim=0cm 0cm 0cm 0cm, clip]{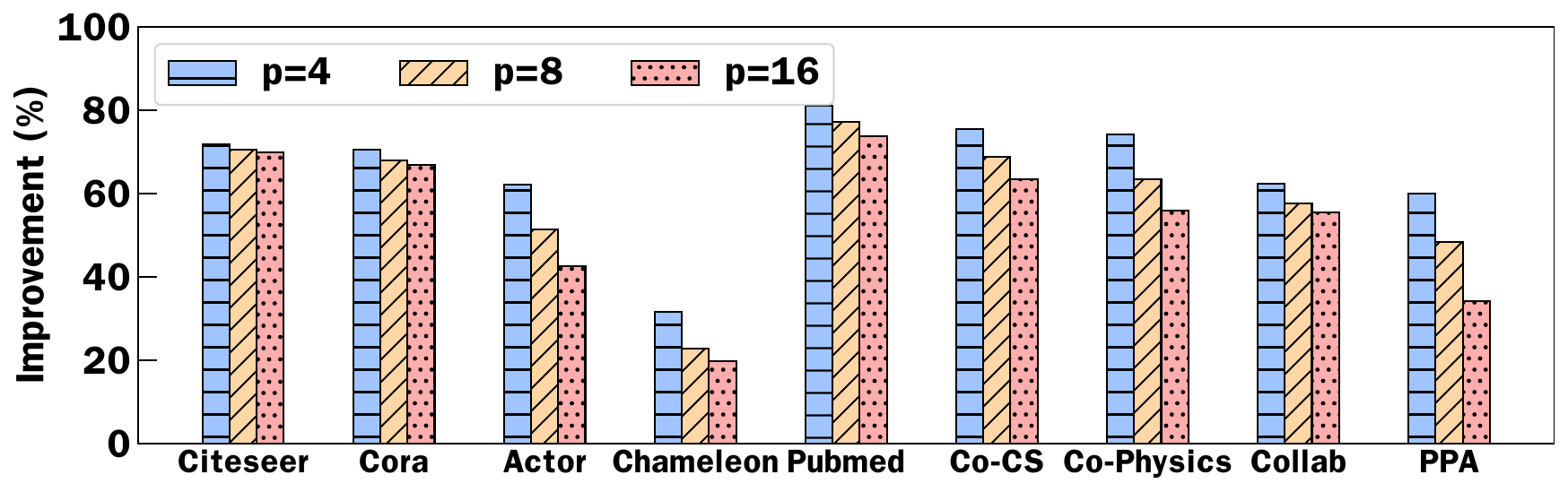}
    }
    \hspace{-2mm}
    \subfloat[\n{} vs. RandomTMA+]{%
    \includegraphics[width=0.325\linewidth, trim=0cm 0cm 0cm 0cm, clip]{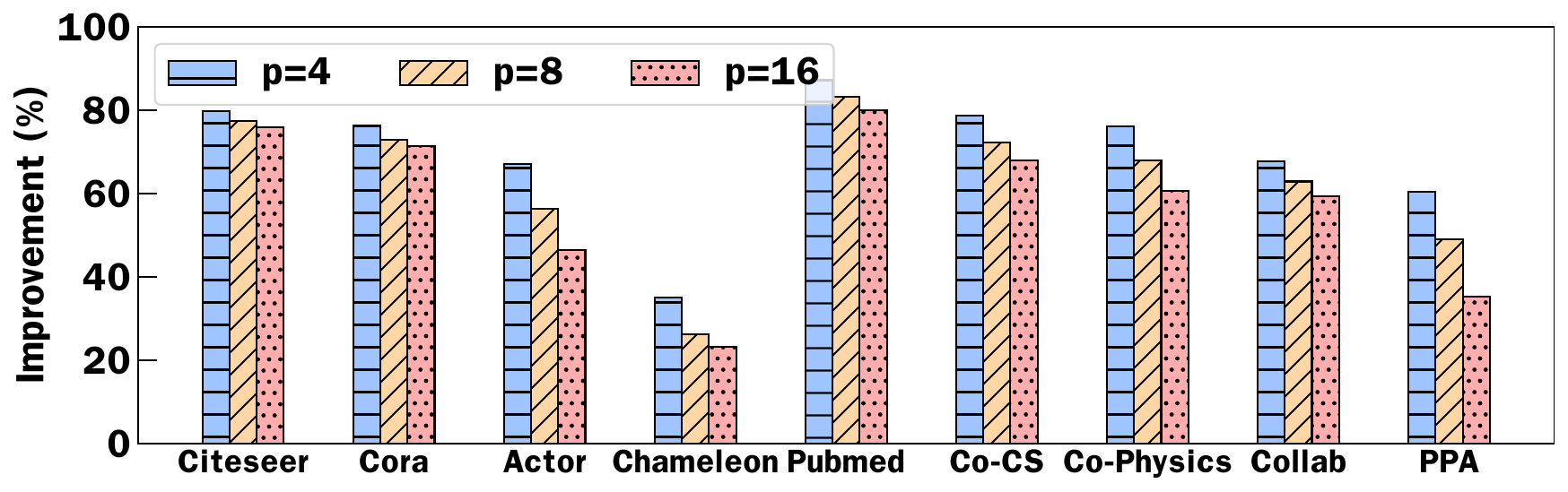}
    }
    \hspace{-2mm}
    \subfloat[\n{} vs. SuperTMA+]{%
    \includegraphics[width=0.325\linewidth, trim=0cm 0cm 0cm 0cm, clip]{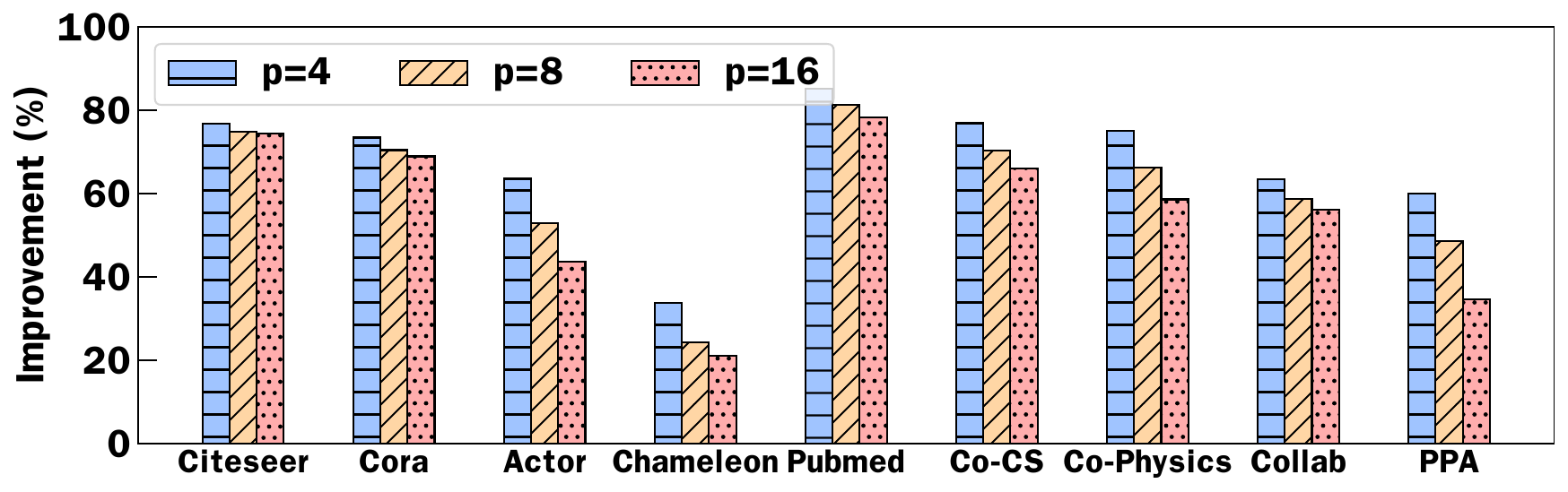}
    }
    \vspace{-1mm}
    \caption{Improvement of communication cost achieved by \n{} for (a)--(c) GCN and (d)--(f) GraphSAGE.}
    \label{fig:sp_improvement}
    \vspace{-2mm}
\end{figure*}

\vspace{2pt}
\noindent \textbf{Hardware and software configuration.} All experiments are conducted on two Linux servers. The first one is equipped with two NVIDIA A6000 48-GB GPUs. The second one is a Lambda Cloud instance with eight NVIDIA V100 16-GB GPUs.\footnote{\url{https://lambdalabs.com/}} For the cases where the number of GPUs is smaller than the number of partitions, we emulate distributed training environments by running multiple processes on each GPU to process multiple subgraphs in parallel. For example, to use two GPUs for four partitions, each GPU processes two partitions to train two local models in parallel. We run the experiments on both servers and confirm that their results are identical. For software, we use Python 3.11, PyTorch v2.0.1, DGL v1.1.0, and CUDA 11.8.

\begin{figure}[t!]
    \captionsetup[subfloat]{captionskip=1pt}
    \centering
    \vspace{-2mm}
    \subfloat[GCN]{%
    \includegraphics[width=0.8\linewidth, trim=0cm 0cm 0cm 0cm, clip]{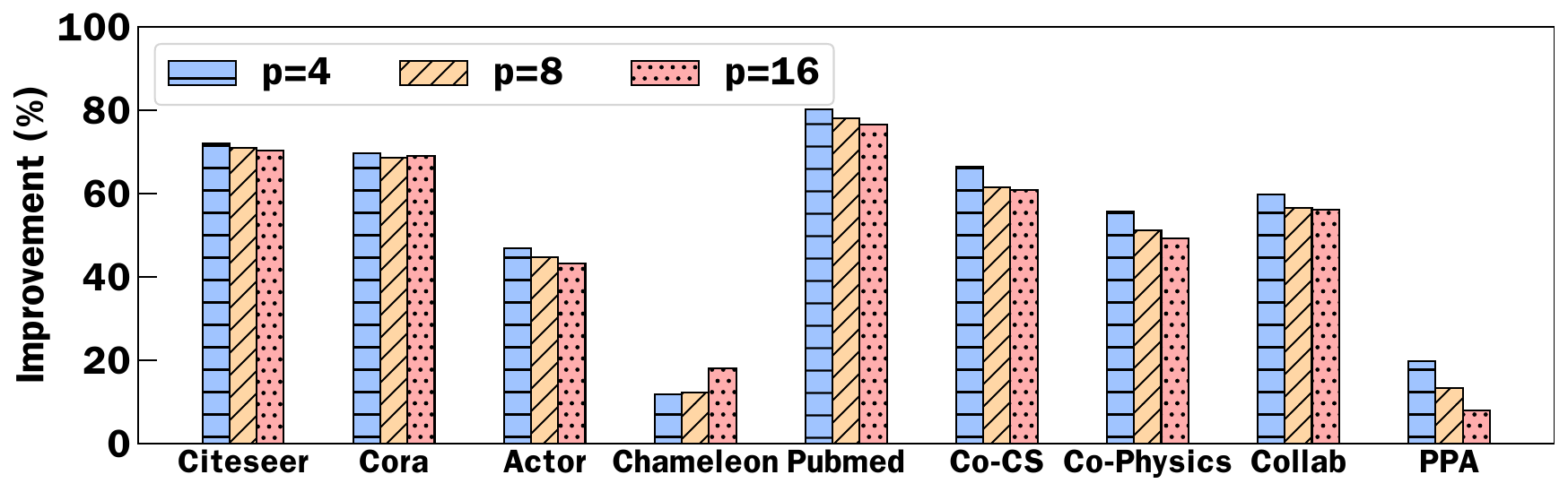}
    }
    \vspace{0mm}
    \subfloat[GraphSAGE]{%
    \includegraphics[width=0.8\linewidth, trim=0cm 0cm 0cm 0cm, clip]{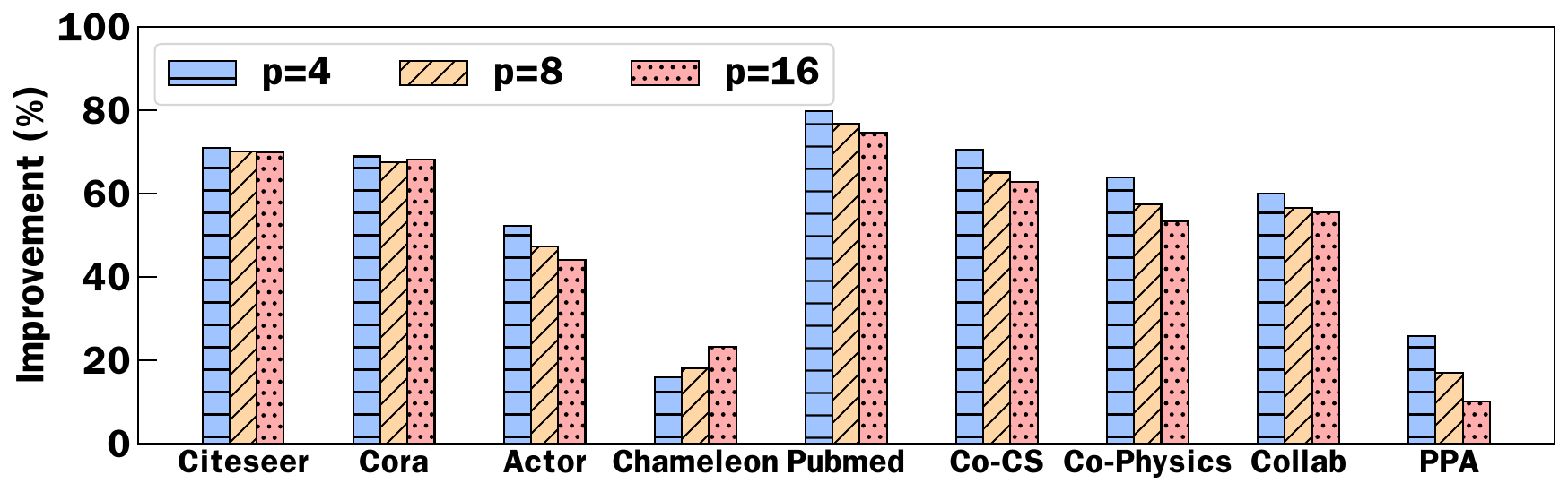}
    }
    \vspace{-1mm}
    \caption{Improvement of communication cost achieved by \n{} over \n{}+.}
    \label{fig:sp_plus_improvement}
    \vspace{0mm}
\end{figure}

\vspace{2pt}
\noindent \textbf{Datasets.} We consider nine public datasets whose statistics are summarized in Table~\ref{table:dataset}. The first seven datasets can be found in DGL~\cite{wang2019dgl} and the last two are from OGB~\cite{hu2020open}. For DGL datasets, we randomly select 80\% edges as training edges, 10\% as validation edges, and the remaining 10\% as test edges. We randomly generate source-destination pairs that are not connected via edges as negative samples (edges). The number of negative samples is the same as that of positive samples for training, and their numbers for validation and testing are three times the corresponding numbers of positive samples for validation and testing, respectively~\cite{wang2019dgl}. For OGB datasets, we follow their split rules.

\vspace{2pt}
\noindent \textbf{Hyperparameters.} We use a 3-layer GCN and a 3-layer GraphSAGE, each with a 3-layer MLP edge predictor, as representative GNN models for link prediction, unless otherwise noted. The hidden dimensions are all 256. For GraphSAGE, we sample 25, 10, and 5 nodes from the first-, second-, and third-hop neighbors, respectively. By default, the models are trained in a mini-batch manner with a batch size of 256 for all DGL datasets. Due to the big sizes of OGB datasets, we use a batch size of 10240 and 51200 for Collab and PPA, respectively. We use the Adam optimizer and set the learning rate to 0.001. We use Hits@100 as the accuracy metric, which is commonly used in the literature~\cite{hu2020open}. We train the model for 500 epochs and report the test accuracy using the model trained with the best validation accuracy. 

All baselines are designed to use model averaging~\cite{li2019convergence} for model synchronization. Thus, we develop \n{} to support both gradient averaging~\cite{li13pytorch} and model averaging and confirm that their prediction performance remains more or less the same. For brevity, here we only report the results obtained under model averaging. We partition the graphs into $p \!\in\! \{4,8,16\}$ partitions for distributed training. For graph sparsification, we set the value of $L^i$ to be proportional to the number of edges in each partition $\mathcal{G}^i$ in order to make the sparsification level \emph{consistent} across datasets, i.e., $L^i \!:=\! \alpha |\mathcal{E}^i|$, where the value of $\alpha$ is controlled to determine the sparsification level. Unless otherwise noted, we set $\alpha\!=\!0.15$, leading to about 85\% edges that are removed after sparsification. The experiments are repeated multiple times to ensure that the results are consistent across different runs.

\begin{table}[t!]
\renewcommand{\arraystretch}{1.1}
\vspace{-4mm}
\caption{Running time of the effective resistance-based graph sparsification of \n{} in seconds}
\vspace{-1mm}
\label{table:time}
\centering
\footnotesize
\begin{adjustbox}{width=0.7\columnwidth,center}
    \begin{tabular}{|c|c|c|c|c|}
    \hline
     & $p\!=\!4$ & $p\!=\!8$ & $p\!=\!16$ \\
    \hline
    \hline
    Citeseer & 2.657 & 2.879 & 2.964 \\
    \hline
    Cora & 2.779 & 2.821 & 2.922 \\
    \hline
    Actor & 3.512 & 3.613 & 3.661 \\
    \hline
    Chameleon & 3.283 & 3.511 & 3.815 \\
    \hline
    Pubmed & 3.465 & 3.769 & 4.201 \\
    \hline
    Co-CS & 4.203 & 4.456 & 4.676 \\
    \hline
    Co-Physics & 7.752 & 8.051 & 8.806 \\
    \hline
    Collab & 32.034 & 32.624 & 32.984 \\
    \hline
    PPA & 571.871 & 581.299 & 595.520 \\
    \hline
    \end{tabular}
\end{adjustbox}
\vspace{0mm}
\end{table}

\begin{figure*}[ht]
    \captionsetup[subfloat]{captionskip=1pt}
    \centering
    \subfloat[\n{} vs. PSGD-PA]{%
    \includegraphics[width=0.325\linewidth, trim=0cm 0cm 0cm 0cm, clip]{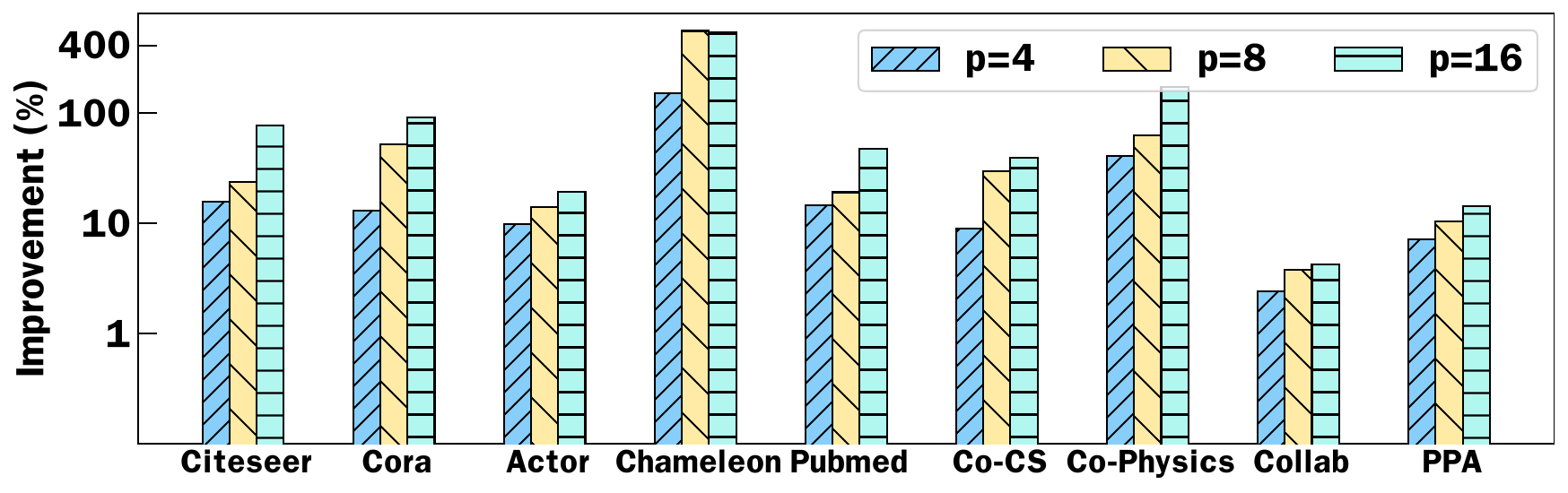}
    }
    \hspace{-2mm}
    \subfloat[\n{} vs. RandomTMA]{%
    \includegraphics[width=0.325\linewidth, trim=0cm 0cm 0cm 0cm, clip]{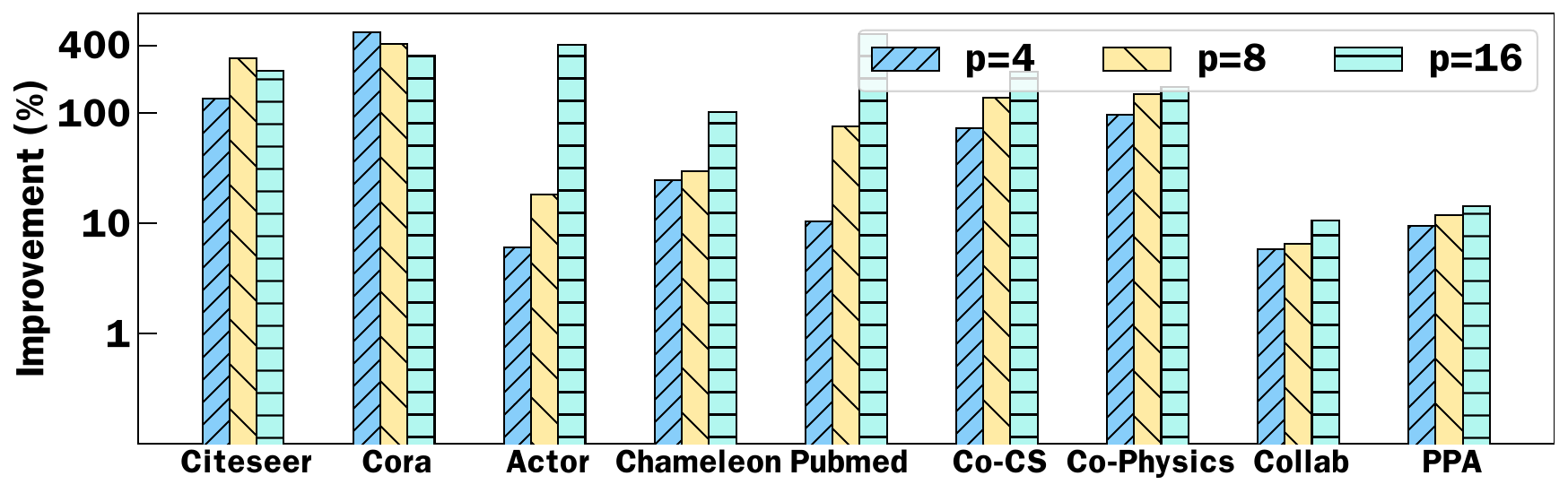}
    }
    \hspace{-2mm}
    \subfloat[\n{} vs. SuperTMA]{%
    \includegraphics[width=0.325\linewidth, trim=0cm 0cm 0cm 0cm, clip]{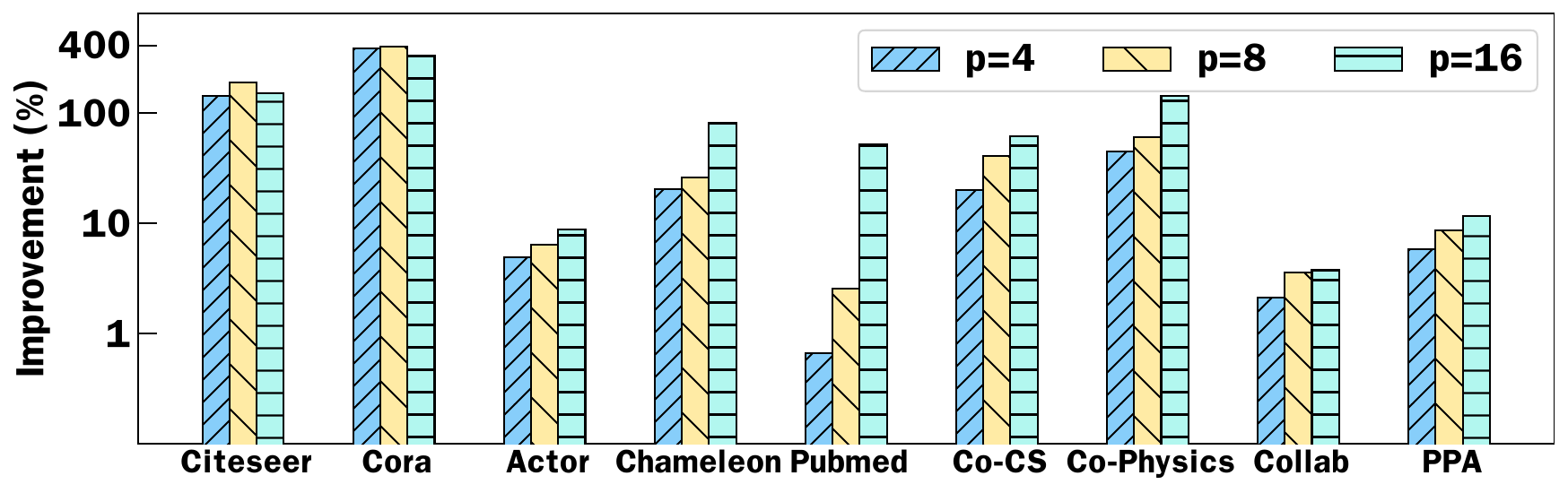}
    }
    \vspace{0mm}
    \subfloat[\n{} vs. PSGD-PA]{%
    \includegraphics[width=0.325\linewidth, trim=0cm 0cm 0cm 0cm, clip]{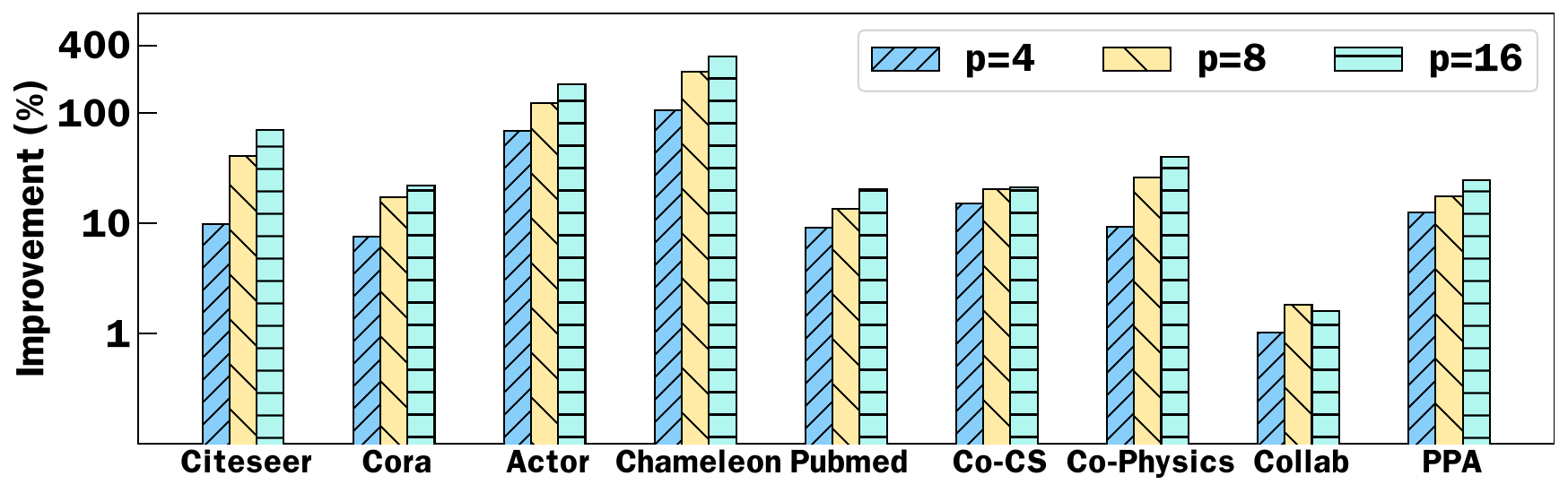}
    }
    \hspace{-2mm}
    \subfloat[\n{} vs. RandomTMA]{%
    \includegraphics[width=0.325\linewidth, trim=0cm 0cm 0cm 0cm, clip]{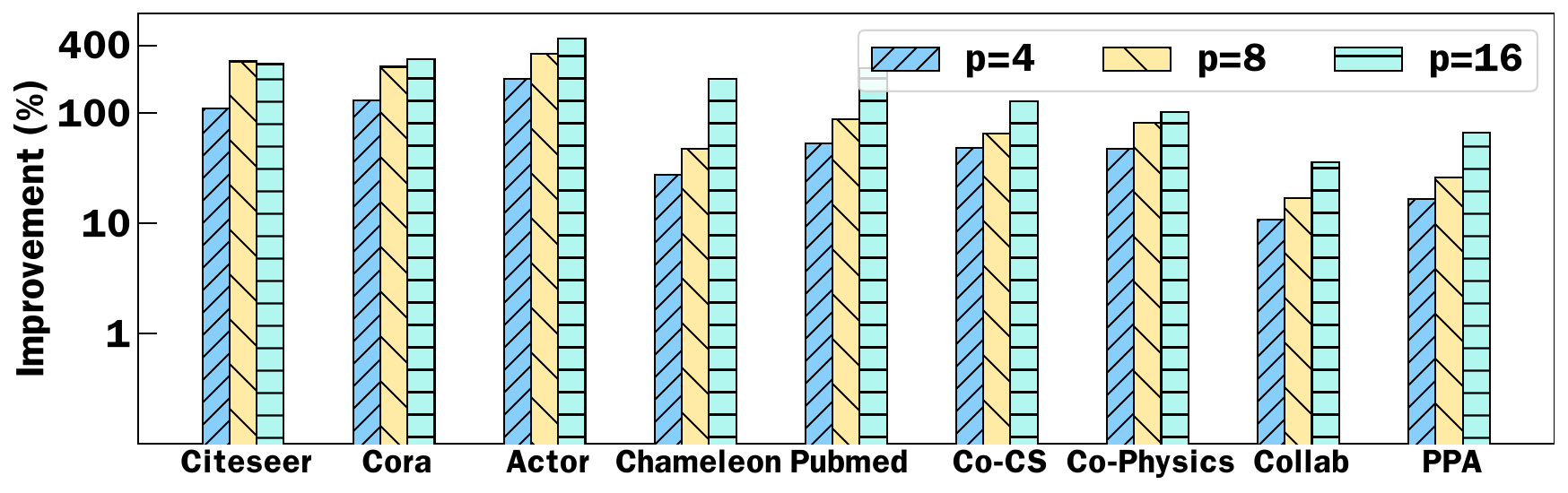}
    }
    \hspace{-2mm}
    \subfloat[\n{} vs. SuperTMA]{%
    \includegraphics[width=0.325\linewidth, trim=0cm 0cm 0cm 0cm, clip]{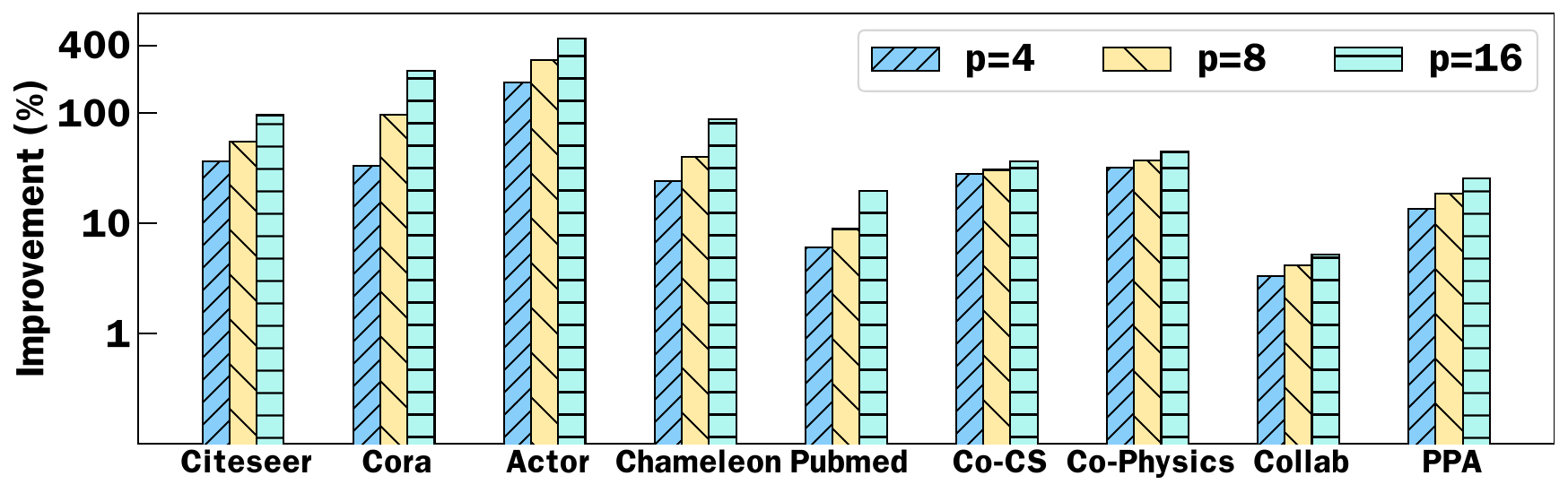}
    }
    \vspace{-1mm}
    \caption{Improvement of accuracy achieved by \n{} for (a)--(c) GCN and (d)--(f) GraphSAGE.}
    \label{fig:sp_improvement_acc}
    \vspace{-2mm}
\end{figure*}

\begin{figure}[t]
    \captionsetup[subfloat]{captionskip=1pt}
    \vspace{-2mm}
    \centering
    \subfloat[GCN]{%
    \includegraphics[width=0.8\linewidth, trim=0cm 0cm 0cm 0cm, clip]{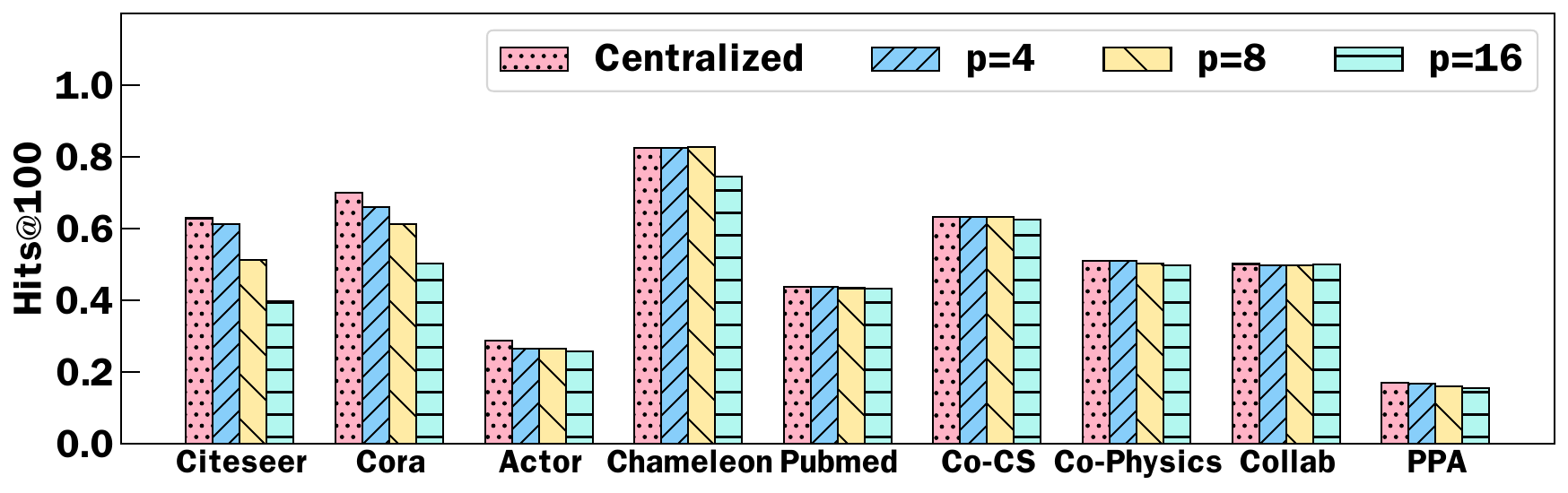}
    }
    \vspace{0.5mm}
    \subfloat[GraphSAGE]{%
    \includegraphics[width=0.8\linewidth, trim=0cm 0cm 0cm 0cm, clip]{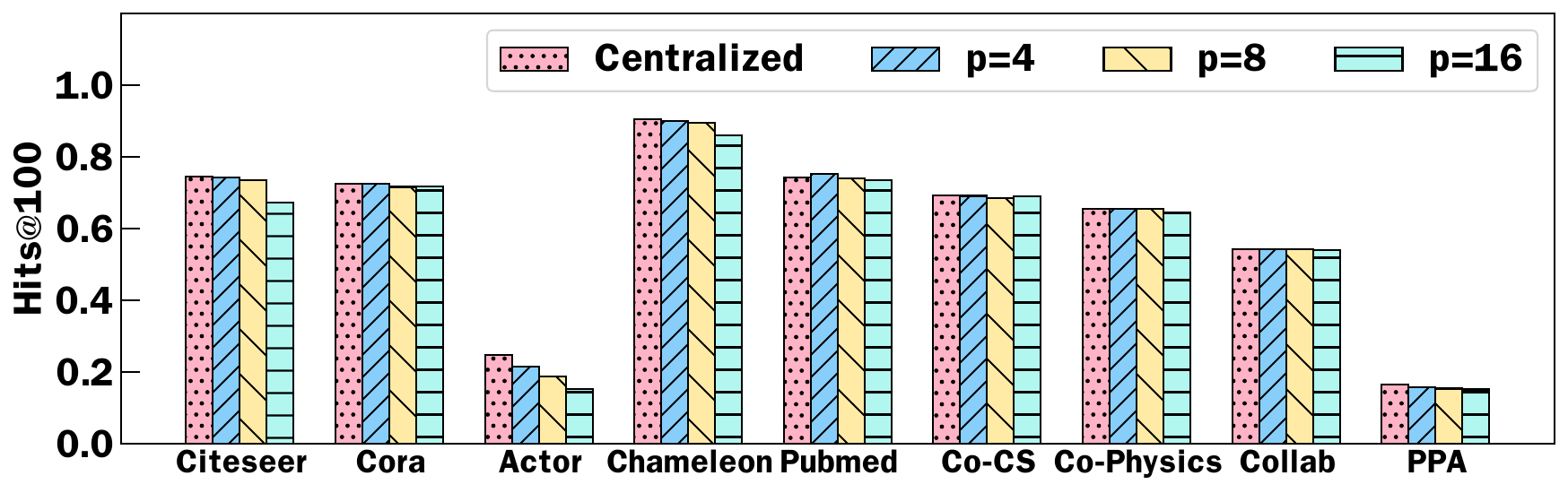}
    }
    \vspace{-0.5mm}
    \caption{Accuracy of GNNs trained by \n{}.}
    \label{fig:sp_acc}
    \vspace{-1mm}
\end{figure}

\subsection{Performance}

\noindent \textbf{Efficiency.} We first demonstrate the efficiency of \n{}. To this end, we measure and compare the communication cost of \n{} against the baselines with the complete data-sharing strategy, i.e., PSGD-PA+, RandomTMA+, and SuperTMA+. We also consider a variant of \n{}, where the complete data-sharing strategy is used (without graph sparsification). We use \n{}+ to denote this variant. We present the improvement of communication cost achieved by \n{} in Figure~\ref{fig:sp_improvement} and Figure~\ref{fig:sp_plus_improvement}. We observe that \n{} can save the communication cost substantially compared to the methods with the complete data-sharing strategy, with an improvement up to about 80\%. It is because the effective resistance-based sparsification algorithm used in \n{} effectively removes unimportant edges in partitioned subgraphs that are used for drawing negative samples. Such sparsed subgraphs lead to a much smaller computational graph (with much fewer $k$-hop neighbors) for the destination node of each global negative sample when it belongs to the sparsed subgraphs. Thus, they highly reduce the amount of data to be transferred during model training. In addition, \n{} maintains the all neighbors (and features) of each node in each partitioned subgraph managed by each worker to eliminate the need for transferring them during model training, which further reduces the communication cost.

\n{} introduces an additional cost due to graph sparsification, but the cost is minimal. To see this, we measure the running time of the graph sparsification algorithm in \n{} and report the results in Table~\ref{table:time}. We observe that the running time of sparsification is just a few seconds for small graphs and a few minutes for large ones under different testing cases. This additional cost is negligible when compared to the overall training time of a GNN model, which typically spans several hours. In other words, the sparsification algorithm in \n{} introduces an insignificant overhead while providing a huge benefit of reducing the communication cost during distributed GNN training for link prediction.

\vspace{2pt}
\noindent \textbf{Effectiveness.} We next demonstrate the effectiveness of \n{} by evaluating the link prediction accuracy of GNNs. As shown in Figure~\ref{fig:sp_acc}, \n{} is able to recover the prediction accuracy achieved by centralized model training for most datasets, although its prediction accuracy falls short for few datasets and testing cases. We also observe that the accuracy of GraphSAGE models is often higher than the one of GCN models. Note that the accuracy of GCN models on small graphs, such as Citeseer and Cora, is a bit worse than the one obtained by centralized training. It is somewhat expected since GCN requires the complete neighbors of each node for feature aggregation. While the graph sparsification algorithm in \n{} tends to remove \emph{unimportant} edges, its impact (information loss) becomes noticeable on small graphs as they do not have many edges in the first place.

We further demonstrate that \n{} outperforms the state-of-the-art baselines significantly. To this end, we compare \n{} with the baselines and report the accuracy improvement achieved by \n{} in Figure~\ref{fig:sp_improvement_acc}. We observe that \n{} outperforms all the baselines and the improvement can be up to about 400\%. It is because \n{} makes each worker to have the all neighbors of each node in its own partitioned subgraph and draws negative samples correctly from the entire sample space. All the above experiment results confirm the high efficiency and effectiveness of \n{} in achieving a good balance between communication overhead and prediction accuracy.

\begin{figure}[t]
    \captionsetup[subfloat]{captionskip=1pt}
    \centering
    \vspace{-4mm}
    \includegraphics[width=0.95\linewidth, trim=0cm 0.2cm 0cm 0cm, clip]{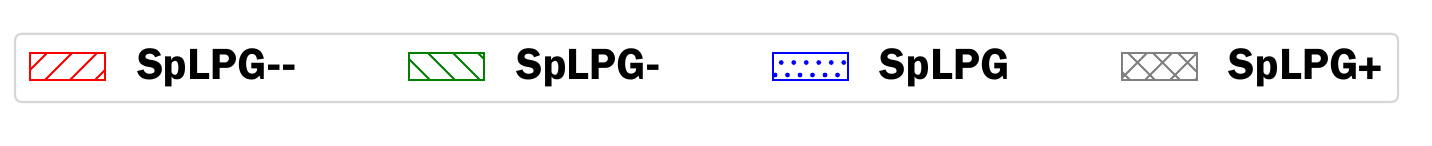}\\
    \vspace{-4.5mm}
    \subfloat[Cora]{%
    \includegraphics[width=0.48\linewidth, trim=0cm 0cm 0cm 0cm, clip]{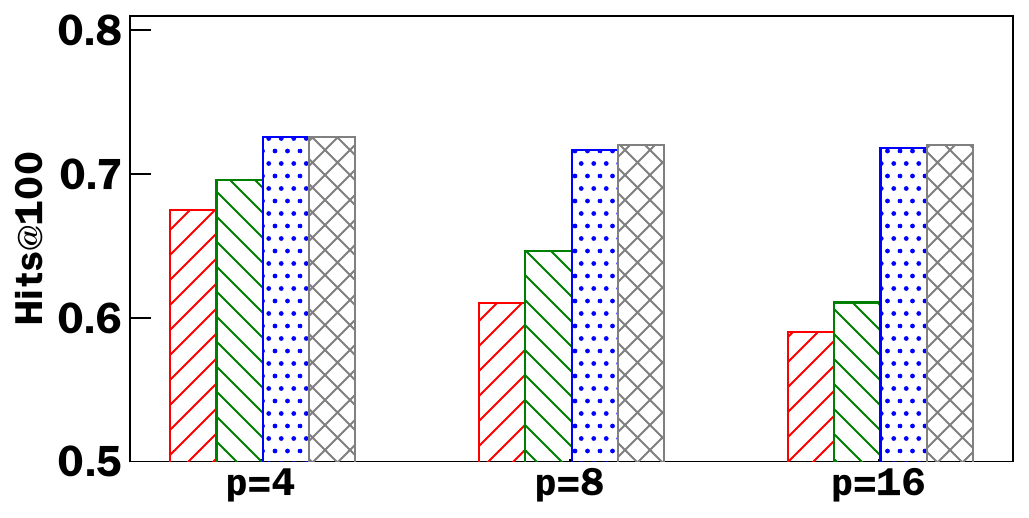}
    }
    \hspace{0mm}
    \subfloat[Chameleon]{%
    \includegraphics[width=0.48\linewidth, trim=0cm 0cm 0cm 0cm, clip]{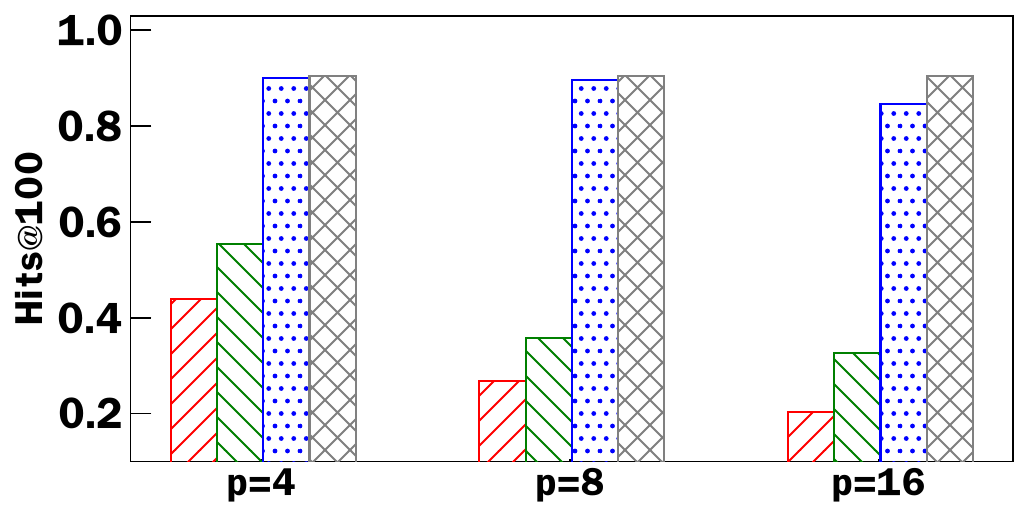}
    }
    \vspace{0mm}
    \subfloat[Pubmed]{%
    \includegraphics[width=0.48\linewidth, trim=0cm 0cm 0cm 0cm, clip]{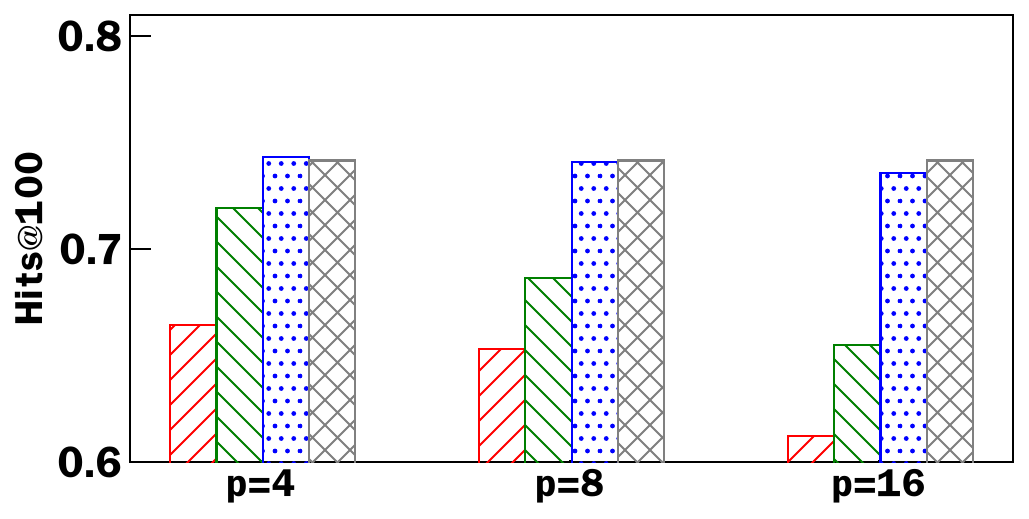}
    }
    \hspace{0mm}
    \subfloat[Co-CS]{%
    \includegraphics[width=0.48\linewidth, trim=0cm 0cm 0cm 0cm, clip]{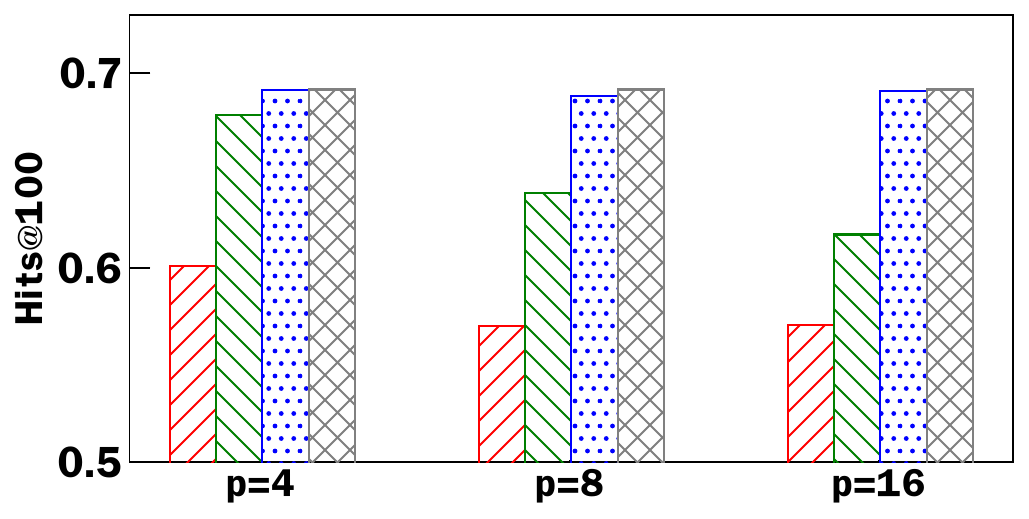}
    }
    \vspace{-0.5mm}
    \caption{Impact of full-neighbors and negative samples.}
    \label{fig:ablation_acc}
    \vspace{-1mm}
\end{figure}

\subsection{Ablation Studies}

We turn our attention to evaluating the impact of the key components and the different choices of hyperparameters of \n{} on the performance. We below report representative results due to the space constraint.

\vspace{2pt}
\noindent \textbf{Impact of full-neighbors and negative samples.} We first study the impact of full-neighbors and negative samples on the performance of GNNs for link prediction. To this end, we consider another two variants of \n{}, named \n{}-{}- and \n{}-. The former does not include the full-neighbors of each node in each partition and is without the data-sharing strategy for drawing global negative samples. The latter includes the full-neighbors of each node in each partition but is without the data-sharing strategy. We report the link prediction accuracy of their corresponding GraphSAGE models along with the ones obtained by \n{} (with sharing sparsed subgraphs) and \n{}+ (with the \emph{complete} data-sharing strategy without graph sparsification) on four datasets as representative examples, as shown in Figure~\ref{fig:ablation_acc}.

We observe that without maintaining the full-neighbors of each node and without drawing global negative samples (i.e., \n{}-{}-), the link prediction accuracy drops substantially. With the full-neighbors maintained but without global negative samples (i.e., \n{}-), the accuracy improves clearly. With both full-neighbor information and global negative samples, \n{} and \n{}+ achieves significant accuracy improvement. The results not only indicate that the information loss that occurs from both graph partitioning and incomplete negative samples is the key source of the performance drop issue in link prediction using distributed GNN training, but also confirm the effectiveness of \n{}.

\begin{figure}[t!]
    \captionsetup[subfloat]{captionskip=1pt}
    \centering
    \subfloat[Communication cost]{%
        \includegraphics[width=0.49\linewidth, trim=0cm 0cm 0cm 0cm, clip]{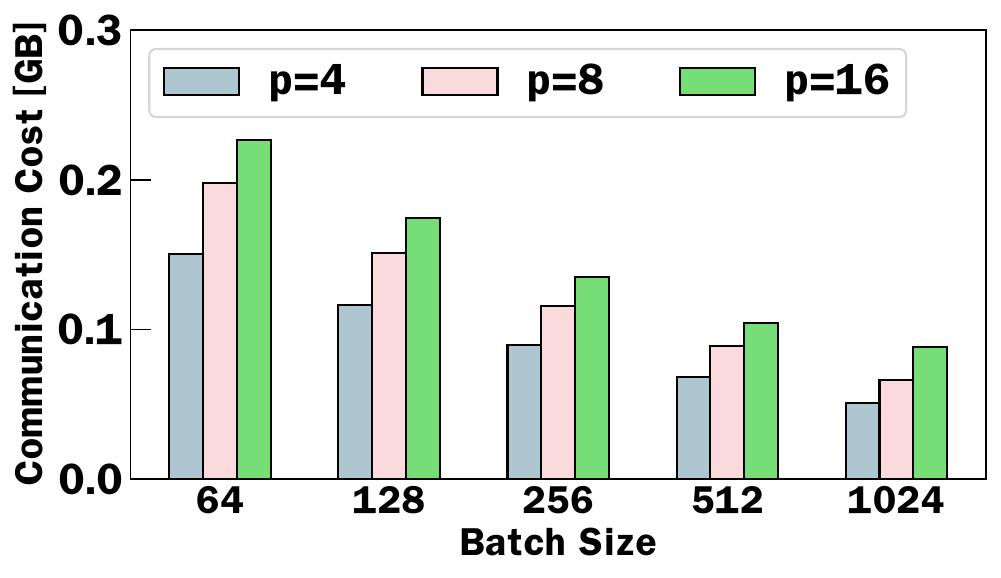}
    }
    \hspace{-2mm}
    \subfloat[Accuracy]{%
        \includegraphics[width=0.49\linewidth, trim=0cm 0cm 0cm 0cm, clip]{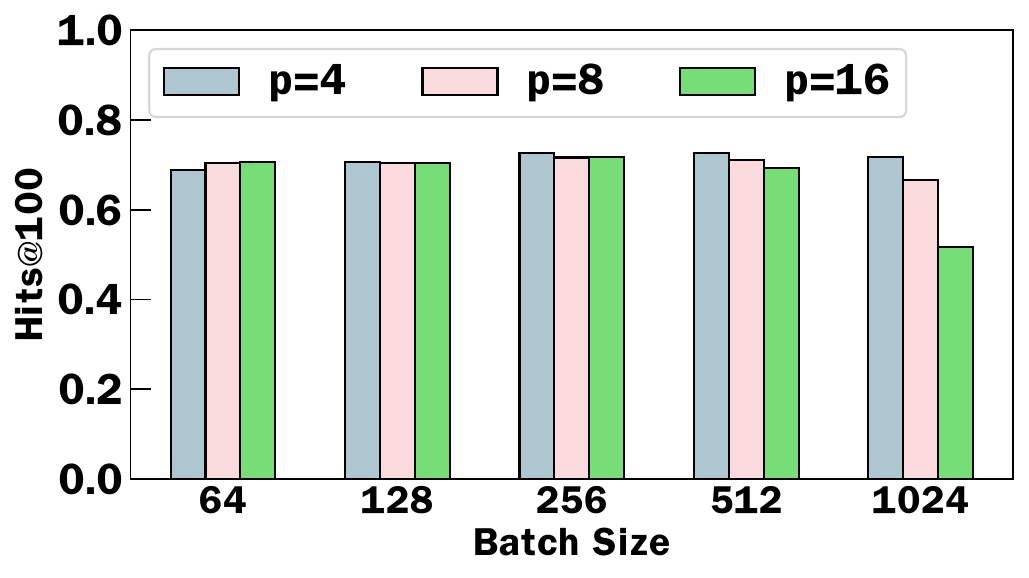}
    }
    \vspace{-1.5mm}
    \caption{Impact of batch size.}
    \label{fig:case_batch}
    \vspace{-3mm}
\end{figure}

\begin{table}[t!]
\renewcommand{\arraystretch}{1.1}
\vspace{-1mm}
\caption{Impact of sparsification level}
\vspace{-1.5mm}
\label{table:Sp_ratio}
\centering
\footnotesize
\begin{adjustbox}{width=0.95\columnwidth,center}
    \begin{tabular}{|c|c|c|c|c|c|c|c|}
    \hline
    & \multicolumn{3}{c|}{Communication cost saving} & \multicolumn{3}{c|}{Accuracy} \\
    \hline
    & $p\!=\!4$ & $p\!=\!8$ & $p\!=\!16$ & $p\!=\!4$ & $p\!=\!8$ & $p\!=\!16$ \\
    \hline
    \hline
    $\alpha=0.05$ & 82.3\% & 82.5\% & 82.9\% & 0.716 & 0.701 & 0.688 \\
    \hline
    $\alpha=0.10$ & 75.5\% & 75.3\% & 75.0\% & 0.722 & 0.713 & 0.711 \\
    \hline
    $\alpha=0.15$ & 68.3\% & 67.6\% & 68.1\% & 0.725 & 0.719 & 0.718 \\
    \hline
    $\alpha=0.20$ & 61.6\% & 62.2\% & 62.5\% & 0.725 & 0.722 & 0.719 \\
    \hline
    \end{tabular}
\end{adjustbox}
\end{table}

\vspace{3pt}
\noindent \textbf{Impact of batch size.} We study the impact of batch size for GNN model training via \n{}. In Figure~\ref{fig:case_batch}, we report the results of GraphSAGE on the Cora dataset as a representative example. We observe that the communication cost of \n{} decreases as the batch size increases. It is because the larger batch size is, the more nodes are in each training batch. As a result, the nodes more likely share common neighbors in the computational graph. As the features of the same node need to be transferred only once per batch, the total cumulative communication cost decreases when batch size increases. We also observe that the accuracy varies slightly for most choices of batch size, but it can drop significantly when batch size is too large. The results indicate that with proper hyperparameter tuning, the communication cost can be further reduced with insignificant prediction accuracy degradation.

\vspace{3pt}
\noindent \textbf{Impact of sparsification level.} We investigate the impact of different sparsification levels on the performance of \n{} and report the results of GraphSAGE on the Cora dataset. The smaller value of $L^i$ for partition $i$ is, the more edges are removed. As presented in Table~\ref{table:Sp_ratio}, we consider $\alpha \!=\! 0.05, 0.10, 0.15$, and $0.20$, which lead to about 95\%, 90\%, 85\%, and 80\% edges that are removed after sparsification, respectively. We report the improvement of communication cost when compared with \n{}+ (without graph sparsification) and the prediction accuracy. We observe that as $\alpha$ grows, the link prediction accuracy increases while the cost saving decreases. $\alpha\!=\!0.15$ turns out to be a reasonable choice to balance the tradeoff between the communication cost and prediction accuracy, as it shows about 68\% cost saving and maintains high accuracy. In other words, it is enough to just maintain about 10\% to 15\% edges in each sparsed subgraph (partition) to achieve high accuracy at a reduced communication cost.

\begin{figure}[t]
    \captionsetup[subfloat]{captionskip=1pt}
    \centering
    \includegraphics[width=\linewidth, trim=0cm 0.2cm 0cm 0cm, clip]{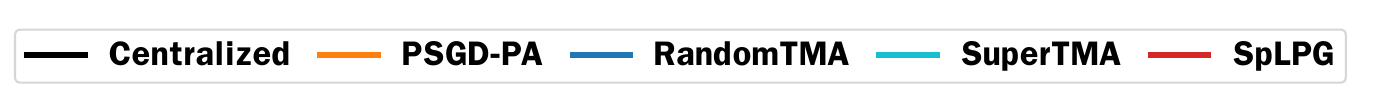}\\
    \vspace{-3.5mm}
    \subfloat[GraphSAGE]{%
        \includegraphics[width=0.48\linewidth, trim=0cm 0cm 0cm 0cm, clip]{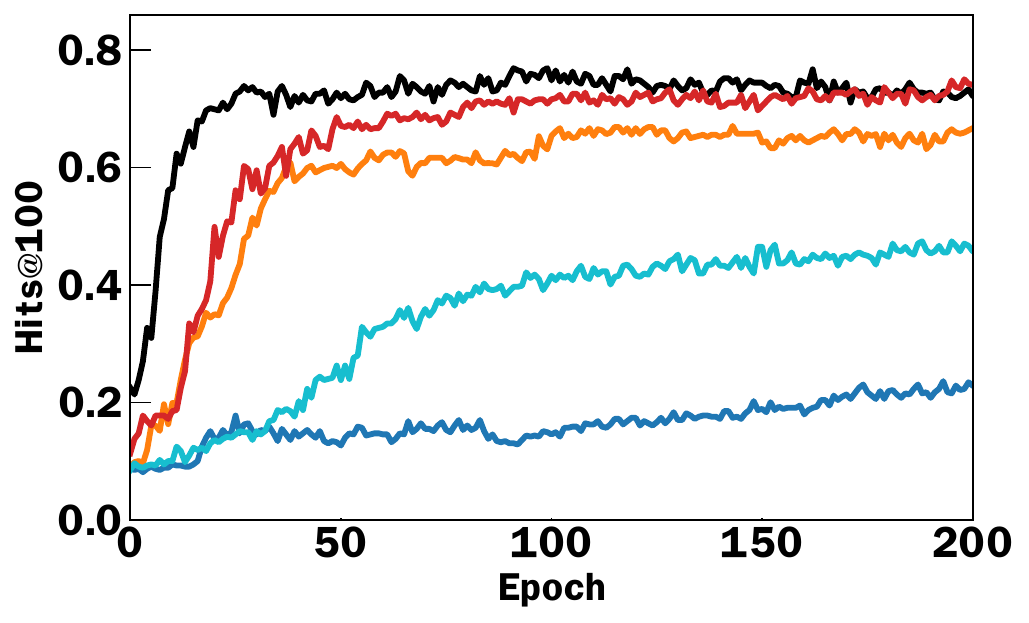}
    }
    \hspace{0mm}
    \subfloat[GCN]{%
        \includegraphics[width=0.48\linewidth, trim=0cm 0cm 0cm 0cm, clip]{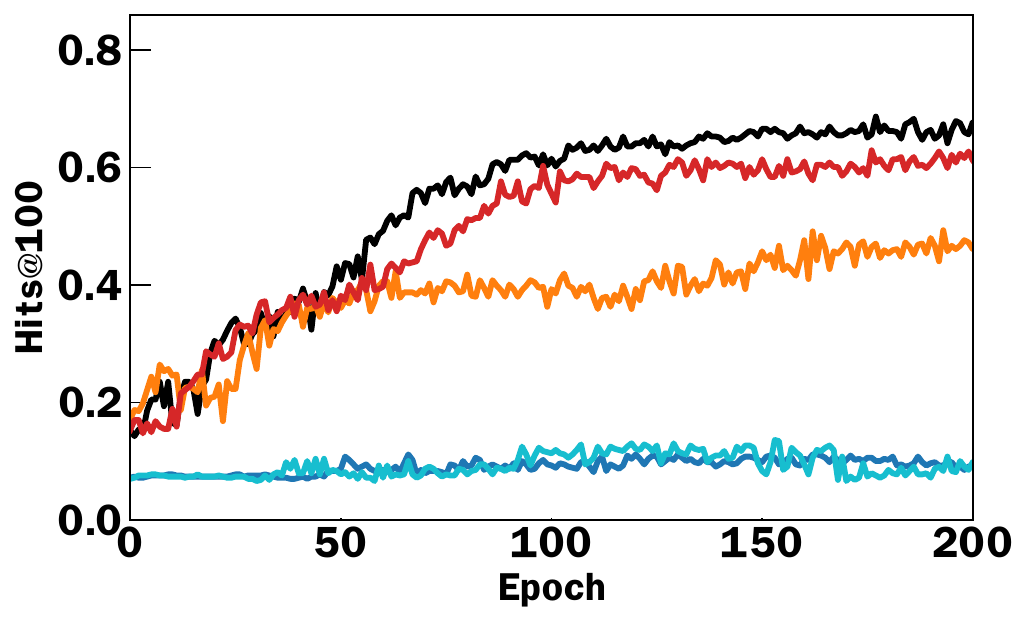}
    }
    \vspace{0mm}
    \subfloat[GAT]{%
        \includegraphics[width=0.48\linewidth, trim=0cm 0cm 0cm 0cm, clip]{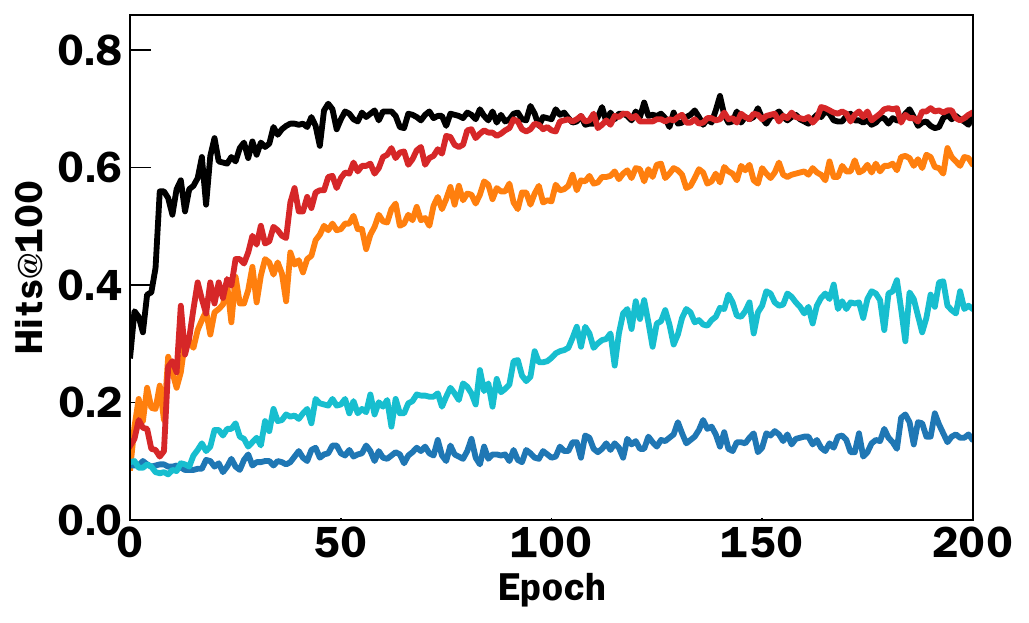}
    }
    \hspace{0mm}
    \subfloat[GATv2]{%
        \includegraphics[width=0.48\linewidth, trim=0cm 0cm 0cm 0cm, clip]{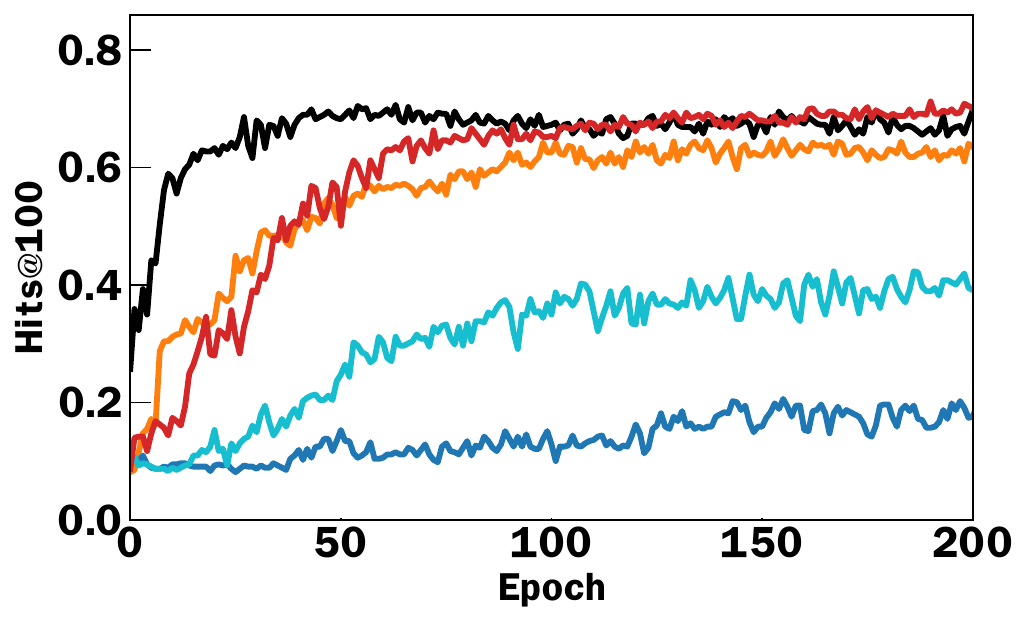}
    }
    \vspace{0mm}
    \subfloat[GraphSAGE]{%
        \includegraphics[width=0.48\linewidth, trim=0cm 0cm 0cm 0cm, clip]{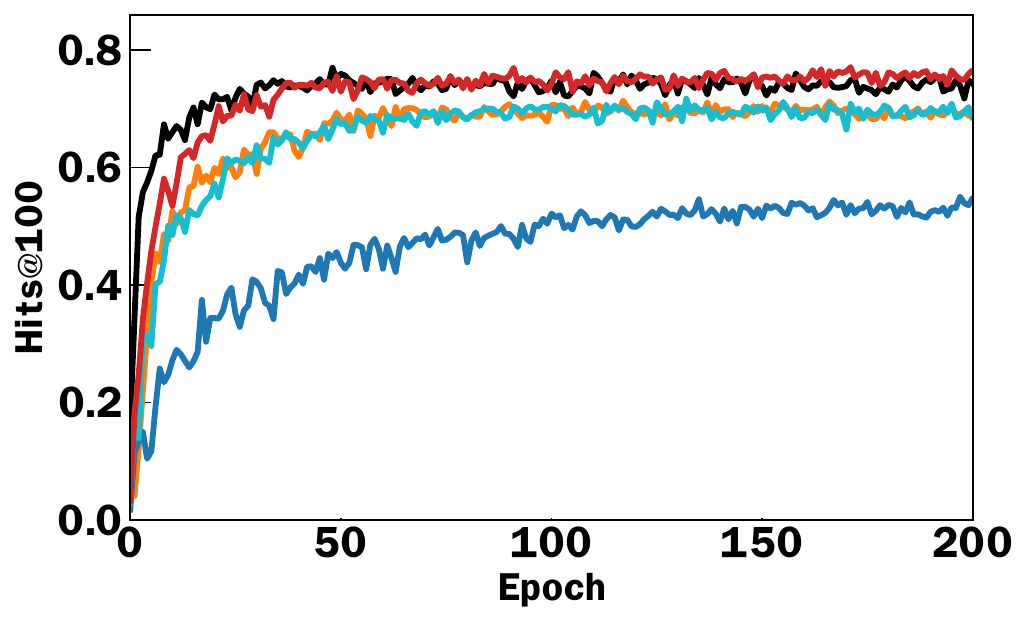}
    }
    \hspace{0mm}
    \subfloat[GCN]{%
        \includegraphics[width=0.48\linewidth, trim=0cm 0cm 0cm 0cm, clip]{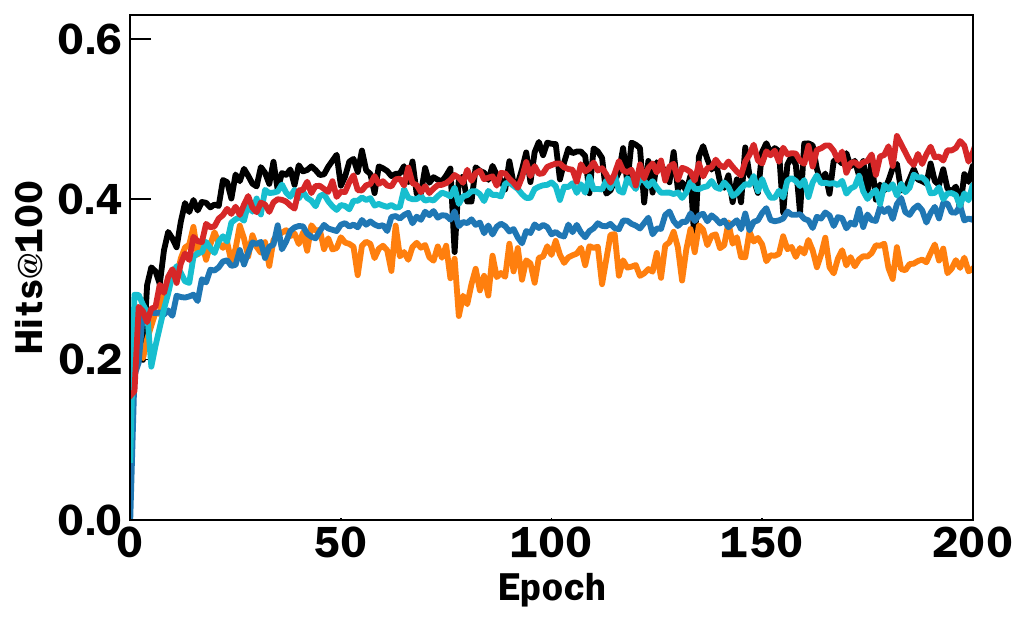}
    }
    \vspace{0mm}
    \subfloat[GAT]{%
        \includegraphics[width=0.48\linewidth, trim=0cm 0cm 0cm 0cm, clip]{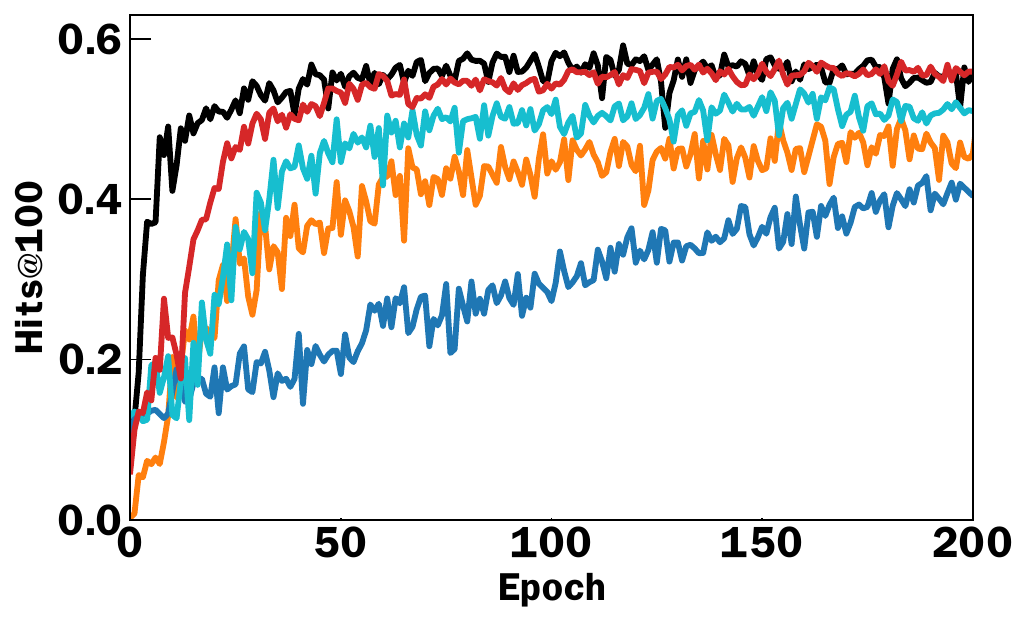}
    }
    \hspace{0mm}
    \subfloat[GATv2]{%
        \includegraphics[width=0.48\linewidth, trim=0cm 0cm 0cm 0cm, clip]{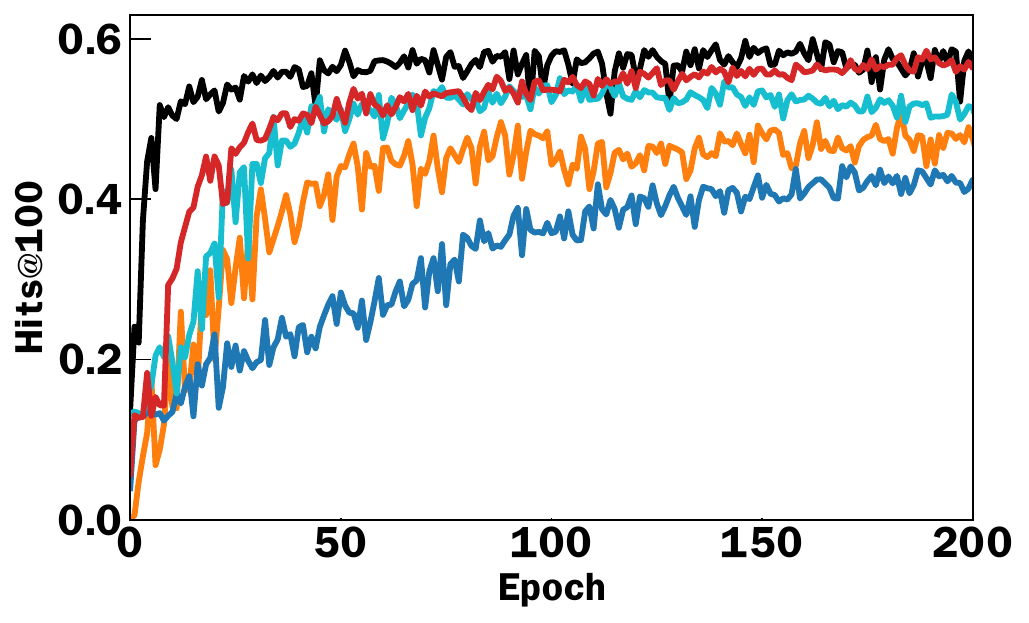}
    }
    \vspace{-0.5mm}
    \caption{Accuracy of different GNN models on (a)--(d) Cora and (e)--(h) Pubmed datasets.}
    \label{fig:models}
    \vspace{-0mm}
\end{figure}

\vspace{3pt}
\noindent \textbf{Different GNN models.} We further test different GNN models and, in Figure~\ref{fig:models}, report the results of GCN~\cite{kipf2016semi}, GAT~\cite{velivckovic2017graph}, and GATv2~\cite{brody2021attentive}, as well as GraphSAGE, trained by \n{} and the baselines. Here we consider the Cora and Pubmed datasets with $p\!=\!4$ as representative examples. We observe that \n{} often converges faster than the baselines and eventually reaches the similar accuracy level to the one achieved under the centralized training. It is because \n{} allows each worker to maintain the full-neighbors of each node in its own partition and correctly draws negative samples from the complete sample space while removing unimportant edges in each subgraph to reduce the communication cost. The results indicate that the performance of \n{} is robust to different GNN models and again confirm the effectiveness of \n{}.

\section{Conclusion}

We have investigated the problem of a performance drop in distributed GNN training for link prediction. We uncovered that its main root causes stem from information loss by graph partitioning and how to draw negative samples in distributed environments. We developed \n{}, which leverages graph sparsification to draw negative samples cost-effectively to strike the right balance between communication cost and prediction accuracy. Extensive experiments demonstrated the superiority of \n{} to state-of-the-art distributed GNN training methods for link prediction.

\section*{Acknowledgment}
This work was supported in part by the National Science Foundation under Grant Nos. 2209921 and 2209922 as well as an equipment donation from NVIDIA Corporation. C.~Lee is the corresponding author.

\bibliographystyle{IEEEtran}
\bibliography{IEEEabrv,ref}

\end{document}